\documentclass{article}

\PassOptionsToPackage{numbers, sort&compress}{natbib}
\usepackage[main, final]{neurips_2026}

\usepackage[utf8]{inputenc} 
\usepackage[T1]{fontenc}    
\usepackage{hyperref}       
\usepackage{url}            
\usepackage{booktabs}       
\usepackage{multirow}       
\usepackage{amsfonts}       
\usepackage{nicefrac}       
\usepackage{microtype}      
\usepackage{xcolor}         
\usepackage{graphicx}       
\usepackage{wrapfig}
\usepackage{xspace}
\usepackage{algorithm}
\usepackage{algpseudocode}
\usepackage{amsmath}
\usepackage{bm}

\usepackage{graphicx}
\usepackage{enumitem}
\usepackage{comment}
\usepackage{makecell}
\usepackage{float} 
\usepackage{marvosym}

\title{\methodname: Focus-Aware Large Avatar Model for One-Shot 4D Animatable Gaussian Head}

\author{
  Yingdong Hu$^{\ast}$\\
  HKUST\\
  Hong Kong SAR, China\\
  \texttt{yhudj@connect.ust.hk} \\
  \And
   Yisheng He$^{\ast}$\textsuperscript{\Letter}\\
  Tongyi Lab, Alibaba Group \\
  Hangzhou, China \\
  \texttt{ethanheysh@gmail.com} \\
  \AND
  Yiming Jiang \\
  Beihang University \\
  Beijing, China \\
  \And
  Zehong Lin \\
  Lingnan University \\
  Hong Kong SAR, China \\
  \And
  Steven Hoi \\
  Tongyi Lab, Alibaba Group \\
  Hangzhou, China \\
  \And
  Jun Zhang \\
  HKUST\\
  Hong Kong SAR, China\\
}

\begin{document}

\newcommand{\methodname}{FA-LAM}

\maketitle
\footnotetext[1]{$^{\ast}$ indicates equal contribution.}
\footnotetext[2]{\textsuperscript{\Letter} Corresponding author. Contact:\texttt{ethanheysh@gmail.com}}

\begin{abstract}
We propose \methodname, a \textbf{F}ocus-\textbf{A}ware \textbf{L}arge \textbf{A}vatar \textbf{M}odel for one-shot animatable Gaussian head creation, while simultaneously enabling static 3D and dynamic 4D full-head recovery. The core of our method lies in a thorough analysis of the attention mechanisms and the entangled reconstruction and animation training pipeline adopted by prior state-of-the-art approaches. Our analysis identifies two main factors that compromise the quality of 3D full-head generation: (1) incorrect and noisy attention activations, and (2) conflicts between the tasks of reconstruction and animation. To address the first issue, we introduce a symmetric and semantic attention regularization strategy that leverages the inherent semantics and structural symmetry of human heads. To disentangle the objectives of reconstruction and animation, we develop a novel dual-phase training pipeline that separates the model's capabilities for large-view hallucination and animation into distinct modules. Moreover, we enhance our model to support multi-view and streaming 4D reconstruction in an efficient and memory-friendly manner through a core autoregressive modification with tailored visibility-aware token fusion. Collectively, these innovations enable \methodname~to reconstruct animatable Gaussian full heads with superior quality, particularly in fine facial regions and large viewing angles.
\end{abstract}

\section{Introduction}
The generation of animatable human avatars from a single image and the reconstruction of dynamic 4D human heads from monocular video streams are fundamental challenges in digital human modeling. 
These tasks are critical for emerging applications, such as virtual try-on, telepresence, AR/VR, and live performance digitization. Nevertheless, it is extremely difficult to solve both tasks simultaneously due to the severe ambiguity inherent in monocular inputs and the need to balance high-fidelity appearance reconstruction and controllable expression animation.

Traditional methods address these tasks by overfitting Gaussian avatars~\cite{gaussianavatars} or neural radiance fields (NeRFs)~\cite{portrait4d, portrait4dv2} to monocular or multi-view videos~\citep{fate, surfhead, gaussianavatars, avatarmav}. While effective, the per-subject optimization is computationally heavy and yields poor novel view synthesis for viewpoints far from the training cameras. Another line of work leverages large multi-view, multi-expression datasets~\cite{xu20243d} to learn strong human priors that enable few-shot inference~\cite{headgap,one2avatar,chen2024monogaussianavatar}. However, the limited model capacity often leads to degraded quality when generalizing to unseen identities or extreme poses.

Recently, feed-forward transformer-based reconstruction models~\citep{lrm, lam, he2026meshlam, panolam, avat3r} have demonstrated compelling inference speed and generalization. They embed head representations, such as FLAME~\cite{flame} tokens, and apply cross-attention to image features to directly regress 3D Gaussian assets, bypassing test-time optimization. Some works~\cite{panolam,forge4d} extend this idea to 4D modeling by learning a direct mapping from images to dynamic representations. Despite promising progress, existing methods exhibit several critical limitations. First, attention mechanisms are operated without explicit regularization on the relationship between avatar tokens and image regions, often ignoring the inherent bilateral symmetry of human heads and causing blurry textures in fine detail areas such as hair. Second, these methods suffer from inherent conflicts between reconstruction fidelity and animation quality, making it difficult to generate avatars that simultaneously achieve high-quality novel view rendering and accurate expression transfer under joint supervision. Third, naively extending such models to multi-view~\cite{vggt} or streaming scenarios~\citep{streamvggt} results in quadratic computational growth with the number of input frames, making them impractical for real-time or resource-constrained settings.

In this work, we present a unified framework to address these challenges with three key innovations. 
First, to address the lack of explicit attention guidance in existing methods, we provide an in-depth analysis of the attention mechanisms within large avatar reconstruction networks. Based on this analysis, we \textbf{propose a semantic and symmetric attention regularization method} that explicitly encourages the model to focus on semantically meaningful and spatially symmetric image regions, leveraging the natural bilateral symmetry of human heads. This regularization significantly improves the clarity of the rendering of reconstructed head avatars, particularly around fine facial features.
Second, to resolve the common frontal-view bias and significant large-view distortion observed in prior works, we identify that the underlying cause is gradient conflict between two supervision objectives: novel-view synthesis (reconstruction) and novel-expression transfer (animation). Through empirical gradient analysis, we show that jointly training with both objectives leads to competing updates that degrade performance, especially for large viewing angles. To resolve this, we \textbf{propose a simple yet effective two-stage training pipeline:} the first stage focuses purely on static 3D reconstruction from multi-view images, while the second stage specializes in refining expression-driven animation only on highly dynamic regions from single-view or multi-view videos. This decoupling enables our model to excel at two distinct tasks within a unified architecture: static full-head 3D reconstruction and animatable avatar creation. Third, to support efficient multi-view and streaming scenarios, we extend our framework to handle sequential input streams via an autoregressive reconstruction design. Specifically, we \textbf{introduce a visibility-gated Gaussian fusion mechanism} that selectively updates the canonical avatar representation based on the visibility map of each incoming frame. This design ensures constant memory consumption regardless of sequence length and allows our model to temporally align 4D Gaussian assets from a monocular stream, updating the head avatar incrementally as new observations arrive. To facilitate effective training of our model, we curate a multi-view extension of the VFHQ dataset~\citep{vfhq}. We augment it with high-quality side-view images generated by Qwen-Image-Edit~\citep{qwenimageedit}, providing stronger supervision for wide viewing angles.

In summary, our contributions are as follows:
\vspace{-3mm}

\begin{itemize}
\item We propose a large avatar reconstruction model for one-shot animatable Gaussian full-head avatars, achieving improved attention activation, enhanced detail reconstruction, and superior novel view synthesis quality. The model is further extended into an autoregressive pipeline with visibility-gated Gaussian fusion, enabling streaming and multi-view reconstruction under constant memory consumption.

\vspace{-1mm}
\item We introduce a semantic and symmetric attention regularization method that directs attention toward semantically meaningful and spatially relevant image regions according to human head symmetry, improving overall reconstruction fidelity and fine-grained rendering.

\vspace{-1mm}
\item We identify that performance degradation in large-view synthesis arises from the gradient entanglement between reconstruction and animation objectives, and resolve it with a dual-phase training strategy that decouples these two learning signals.
\end{itemize}
\section{Related Work}
\subsection{Images to Head Avatars}
Early 2D-based methods for image-to-head avatar reconstruction typically rely on CNNs and GANs~\citep{goodfellow2014generative} to synthesize morphable faces. Some inject pose and expression features into generators~\citep{burkov2020neural, wang2023progressive, zakharov2019few, hong2022depth}, often structured as UNet or StyleGAN~\citep{karras2019style}. Others warp expressions and head poses as motion fields on source views~\citep{drobyshev2022megaportraits, guo2024liveportrait, zhang2023metaportrait}. Recent efforts have turned to diffusion models for high-quality face generation~\citep{tian2024emo, cui2024hallo2, xu2024hallo}. Notably, CAP4D~\citep{cap4d} and MVP4D~\citep{mvp4d} incorporate 3D Morphable Models (3DMMs)~\citep{flame, loper2023smpl, SMPL-X:2019} to guide novel view synthesis with novel expressions. Despite their strengths, these 2D models suffer from multi-view consistency issues and inefficient inference.
Another line of research lifts images to 3D head avatars, starting with animatable mesh- or point-cloud-based avatars from image sequences~\citep{flame, fate}. Subsequent work have shifted to more powerful representations, including NeRFs~\citep{nerf, portrait4d, portrait4dv2, chu2024gpavatar, kirschstein2024gghead, 10203662} and 3DGS~\citep{3dgs, gaussianavatars, gaussianheadavatar}. Recent works~\citep{soap, headgap} reduce input requirements by leveraging diffusion models for novel view augmentation and reconstruct high-quality avatars from augmented images. However, all these methods require costly iterative optimization per case, lacking the temporal efficiency of feed-forward reconstruction.

\subsection{Feed-Forward 3D Head Avatar Reconstruction}
Feed-forward reconstruction~\citep{lrm, vggt, lam, forge4d} has recently gained attention for its efficiency in unconstrained scenarios, enabled by transformer scaling. In head avatar modeling, recent works~\citep{lam, he2026meshlam, panolam,gagavatar} incorporate human geometry priors into transformers for efficient feed-forward reconstruction. Some approaches~\citep{panolam, gpsgs, evags} reconstruct static representations from inputs, achieving high-quality reconstruction and novel view synthesis, but cannot animate novel expressions or poses. 
Conversely, other works~\cite{lam,he2026meshlam,avat3r,flexavatarpeng,flexavatartobias} introduce parametric priors like FLAME~\citep{flame} into feed-forward frameworks for animatable avatar reconstruction from single or multi-view images. Specifically, LAM~\citep{lam} and MeshLAM~\citep{he2026meshlam} use FLAME priors with transformer attention to reconstruct animatable Gaussians and meshes from a single image. Avat3r~\citep{avat3r} employs transformers as an expression driver to animate DUSt3R~\citep{dust3r} outputs. Concurrent works improve expression driving with transformers~\citep{flexavatartobias} or UNet~\cite{flexavatarpeng}, enhance multi-view capability with large-scale datasets~\cite{flexavatarpeng, uika}, support multi-view inputs~\cite{uika}, and utilize UV space~\cite{uika,bringingportrait}.
However, these methods neglect bilateral facial symmetry and the conflict between reconstruction and animation, leading to poor rendering quality under novel expressions and views. In contrast, we systematically examine two common issues in LRM-based avatar models: detail blurring and large-view degradation. To address them, we propose bilateral semantic attention and a dual-phase training strategy. Furthermore, to handle out-of-memory issues with multiple or streaming frames, we introduce a visibility-aware autoregressive approach.

\begin{figure*}[tb!]
    \centering
    \includegraphics[width=0.88\textwidth]{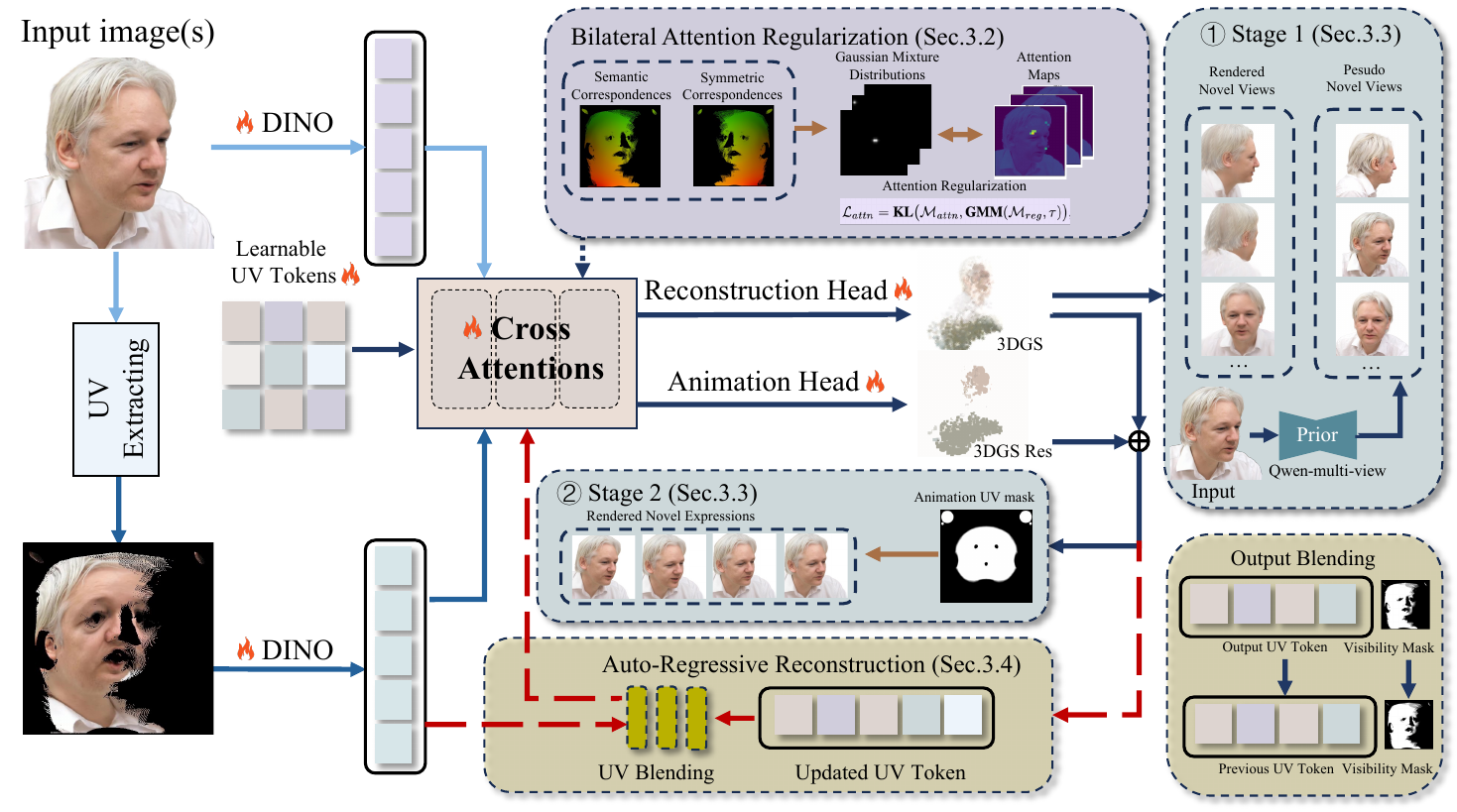}
    \vspace{-4.5mm}
    \caption{\textbf{Overview of \methodname.} Given an input RGB image, we first perform learning‑free coarse UV extraction to obtain a coarse RGB UV map. The UV map and the original image are separately encoded by two DINO encoders. A learnable UV token then queries cross‑attention over both UV and RGB features. The resulting UV tokens are used to decode canonical 3D Gaussians and animation‑specific Gaussian residuals.}
    \vspace{-6mm}
    \label{fig:pipeline}
\end{figure*}
\section{Method}

Given a single portrait image $\mathbf{I}_{in}$, \textbf{\methodname}~aims to construct an animatable 3D Gaussian asset $\mathcal{G}_{3D} = \{\bm{\mu}_k, \bm{s}_k, \bm{r}_k, o_k, \bm{c}_k\}_{k=1}^{M}$ that can be controlled via expression codes $\mathbf{z}_{exp}$ and rendered from arbitrary viewpoints with minimal image distortion. This process can be formalized as:
\begin{equation}
\setlength\abovedisplayskip{1pt}
\mathcal{G}_{3D} = \mathbf{D}_{\theta_2}(\mathbf{E}_{\theta_1}(\mathbf{I}_{in}), \mathbf{T}_{UV}, \bm{\phi}_{FLAME}),
\setlength\belowdisplayskip{1pt}
\end{equation}
where $\mathbf{E}_{\theta_1}$ denotes a pretrained Vision Transformer (ViT) network that extracts features from the input image $\mathbf{I}_{in}$, $\mathbf{T}_{UV}$ is a learnable token map defined in UV space, and $\mathbf{D_{\theta_2}}$ represents a transformer-based decoder that maps learnable tokens to 3D Gaussian attributes. $\bm{\phi}_{FLAME}$ refers to the parametric 3D morphable model (FLAME~\citep{flame}).

In the following, we first introduce the architecture of the encoding and decoding networks in Sec.~\ref{encoderdecoder}. We then detail our proposed attention regularization strategy
in Sec.~\ref{regularization}. Finally, we present a novel two-stage training pipeline in Sec.~\ref{twostage} to disentangle the reconstruction and animation objectives.

\subsection{UV Space Encoding and Decoding}
\label{encoderdecoder}
To improve spatial reasoning of 3D Gaussian tokens, we perform reconstruction in the UV space, as illustrated in Fig.~\ref{fig:pipeline}. Given an input RGB image $\mathbf{I}_{in}$, \methodname~ first extracts a coarse RGB UV image $\mathbf{I}_{UV}$ using a FLAME model animated by the tracked pose, expression, and camera parameters. The extracted UV image $\mathbf{I}_{UV}$, together with the original RGB image $\mathbf{I}_{in}$, are passed through separate DINO encoders $ \mathbf{DINO}_{\theta_{in}}$, $\mathbf{DINO}_{\theta_{UV}}$ to extract image features and UV features, respectively. This process can be formulated as:
\begin{equation}
\setlength\abovedisplayskip{1pt}
    \mathbf{F}_{in}, \mathbf{F}_{UV} = \mathbf{DINO}_{\theta_{in}}(\mathbf{I}_{in}),\; \mathbf{DINO}_{\theta_{UV}}\bigl(\textbf{ExtractUV}(\mathbf{I}_{in})\bigr),
\setlength\belowdisplayskip{1pt}
\end{equation}
where the UV extraction operation \textbf{ExtractUV} is detailed in the Appendix.

Concurrently, a token map $\mathbf{T}_{UV}$ defined in UV space is initialized with learned semantic embeddings and undergoes cross-attention with the extracted image and UV features:
\begin{equation}
\setlength\abovedisplayskip{1pt}
    \mathbf{T}_{UV}^{'} = \textbf{CrossAttention}\bigl(q=\mathbf{T}_{UV},\; kv=\mathbf{F}_{in}\oplus\mathbf{F}_{UV}\bigr).
\setlength\belowdisplayskip{1pt}
\end{equation}
The resulting UV tokens are then passed to a 3D Gaussian decoder, which decodes the semantic 3D tokens into a set of 3D Gaussians as \(\mathcal{G}_{3D} = \mathbf{D}_{3DGS}(\mathbf{T}_{UV}^{'})\).

\subsection{Semantic Attention Regularization with Human Symmetric Prior}
\label{regularization}
\begin{figure}[tb!]
    \centering
    \vspace{-3mm}
    \includegraphics[width=0.8\linewidth]{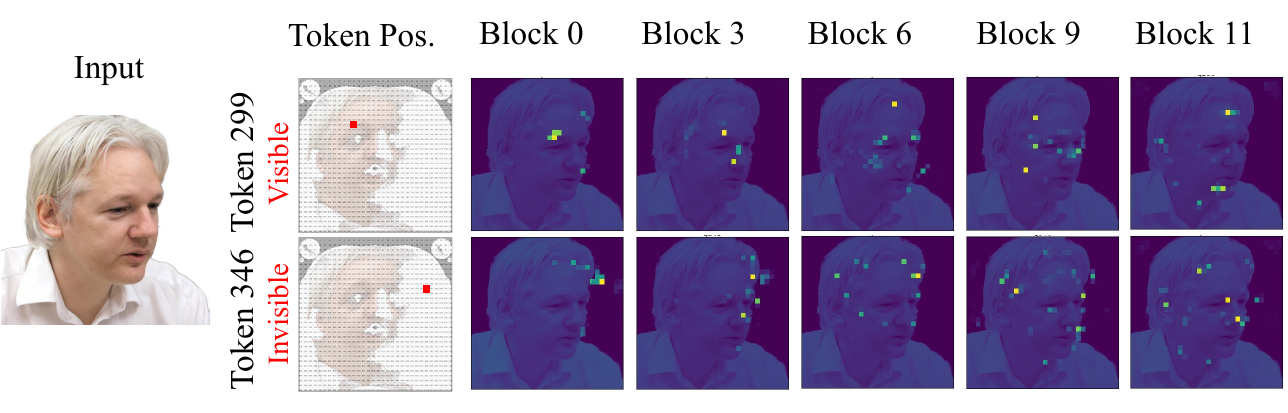}
    \vspace{-4mm}
    \caption{\textbf{Attention map activations} in our network w/o regularization. Token 299 (visible in the input RGB image) exhibits attention mostly on relevant facial regions with some noise, whereas token 346 (not visible) produces largely random activations without coherent semantic focus. This motivates our attention regularization method to improve attention utilization. Please zoom in for detailed views and corresponding UV map token positions.}
    \vspace{-6mm}
    \label{fig:attn}
\end{figure}

A limitation of the vanilla model in Sec.~\ref{encoderdecoder} is that its cross-attention is not spatially well constrained even when trained with large-scale data. To demonstrate this, we train our framework on a large-scale dataset as in LAM~\cite{lam}. We then visualize the activation to analyze how the attention mechanism works. As shown in Fig.~\ref{fig:attn}, UV tokens corresponding to visible facial regions usually activate around semantically relevant image patches, but the responses remain noisy and spatially dispersed. The failure is more severe for tokens corresponding to invisible regions, whose attention often collapses onto irrelevant facial parts or even background clutter. As a result, the model uses its attention budget inefficiently, which degrades reconstruction quality, especially for unseen regions.

This behavior stems from an ambiguity inherent to single-view reconstruction. For visible regions, the model receives direct evidence but still lacks an explicit constraint on where attention should concentrate. For invisible regions, the ambiguity is even worse as no direct image evidence exists, making the learned attention prone to drifting toward spurious but locally plausible patterns. In other words, standard end-to-end training does not provide sufficient structures for the model to distinguish informative correspondences from arbitrary activations.

We address this with semantic attention regularization using a human symmetry prior. Human heads are approximately bilaterally symmetric, especially in UV space, where facial layout is canonically aligned. When one side of the face is occluded, its appearance can often be inferred from the opposite symmetric region. We inject this prior directly into cross-attention layers. For each UV token, we determine its semantic correspondence in the input RGB image by reprojecting the tracked FLAME template into the image plane. Its correspondence in the extracted UV image is obtained by matching the token to the UV patch at the same spatial location. These correspondences provide explicit supervision on attention targets. To handle self-occlusion, we also define symmetric correspondences in both the RGB image and UV map, encouraging tokens of invisible regions to attend to their visible symmetric counterparts instead of drifting.

We implement this constraint by matching the predicted attention map to a target Gaussian mixture distribution centered at the semantic and symmetric correspondences. The regularization loss is
\begin{equation}
\setlength\abovedisplayskip{1pt}
    \mathcal{L}_{attn} = \textbf{KL}\bigl(\mathcal{M}_{attn}, \textbf{GMM}(\mathcal{M}_{reg}, \tau)\bigr),
\setlength\belowdisplayskip{1pt}
\end{equation}
where $\mathcal{M}_{attn}$ denotes the predicted attention map, $\mathcal{M}_{reg}$ denotes the target correspondence locations, $\textbf{GMM}(\cdot,\tau)$ converts these sparse locations into a Gaussian mixture with fixed standard deviation $\tau$, and \textbf{KL} denotes the Kullback--Leibler divergence. This regularization encourages attention to be spatially concentrated, semantically aligned, and symmetry-aware.

\subsection{Disentangling Reconstruction and Animation}
\label{twostage}
Direct training on monocular talking head videos (LAM~\cite{lam}) yields strong animation but poor generalization to unseen views. In contrast, training on static multi-view images (PanoLAM~\cite{panolam}) enables high-quality wide-view reconstruction but cannot animate new expressions. A naive combination of both objectives produces blurry results (Fig.~\ref{fig:abl_2stages}), indicating a negative correlation between reconstruction and animation. To investigate this, we analyze the gradient cosine similarity and the t-SNE feature distributions following PCGrad~\citep{pcgrad} and t-SNE~\citep{tsne}, comparing novel expression and novel view gradients, and visualizing features from networks trained with each objective separately.

\begin{wrapfigure}[15]{r}{0.49\textwidth}
\vspace{-5.8mm}
    \centering
    \includegraphics[width=\linewidth]{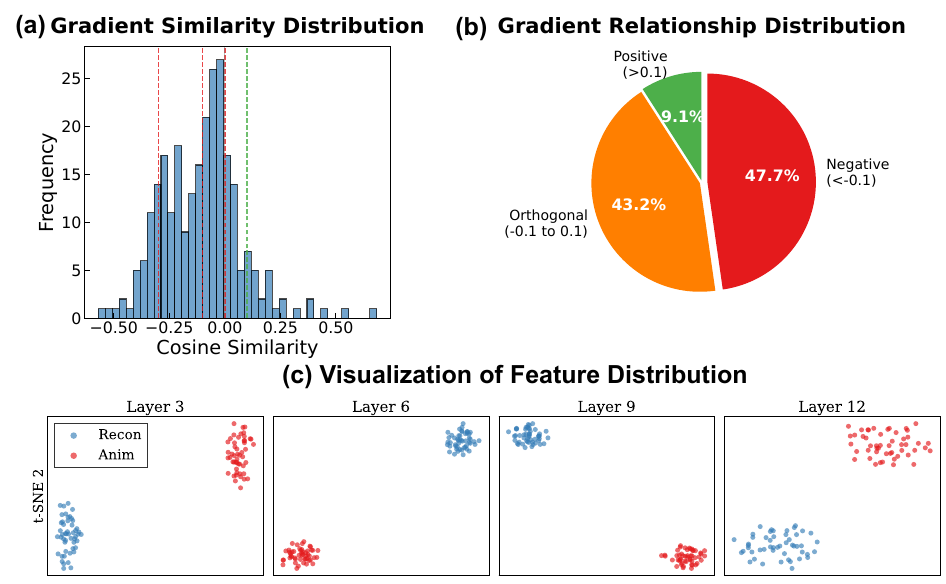}
    \vspace{-7mm}
    \caption{\textbf{Gradient and feature analysis} reveals strong negative correlation between reconstruction and animation objectives, causing performance degradation under joint training.}
    \label{fig:entangle}
\end{wrapfigure} 
As shown in Fig.~\ref{fig:entangle}, attention layer gradients exhibit negative correlation (cosine similarity $<-0.1$) for nearly half of the pairs, with the remainder nearly orthogonal (cosine similarity between $-0.1$ and $0.1$). Such negative correlations hinder multi-task optimization by causing one task to dominate training~\citep{pcgrad}.
To address this issue, we thus propose a two-stage strategy. First, we train for reconstruction only on multi-view portrait data, including our augmented VFHQ dataset, to learn a canonical neutral full-head Gaussian representation with strong cross-view consistency. Second, we fine-tune on animation data for controllable expression dynamics and long-sequence motion. To preserve reconstruction quality, expression deformations are modeled as residual corrections on the canonical Gaussians (Fig.~\ref{fig:pipeline}). Specifically, we predict residual Gaussian attributes only for regions that require large non-rigid motion (periocular and perioral areas), decoded from the same UV tokens via an additional DPT head. This design separates identity-preserving canonical geometry from expression-specific deformation, improving both animation quality and large-view synthesis.

\subsection{Autoregressive Reconstruction with Visibility-Gated Feature Fusion}

\label{ar}We further extend our framework to multi-view and sequential inputs in an autoregressive manner. Unlike full-attention designs, whose memory footprint grows quadratically with the number of input images, our autoregressive model stores only current features and an accumulated state, achieving constant memory per additional frame. The key challenge is that each frame provides only partial evidence: some regions are newly revealed, while others are occluded or unreliable. Directly fusing the accumulated state with incoming predictions thus tends to overwrite stable geometry with inferior observations. To address this, we introduce \emph{visibility-gated feature fusion}, which updates each UV-space Gaussian feature according to its reliability in the current frame.

Given a sequence $\{\mathbf{I}_{in}^{(1)}, \ldots, \mathbf{I}_{in}^{(N)}\}$, \methodname~processes the first frame $\mathbf{I}_{in}^{(1)}$ (Sec.~\ref{encoderdecoder}) to obtain UV tokens $\mathbf{T}_{UV}^{(1)}$ and Gaussian representation $\mathcal{G}_{3D}^{(1)}$. For each $t \geq 2$, we update the latent UV state by fusing the previous tokens with the current UV features:
\begin{equation}
\setlength\abovedisplayskip{1pt}
\mathbf{F}_{UV}^{'(t)} = \textbf{MLP}_{\phi}\left( \mathbf{T}_{UV}^{'(t-1)} \oplus \mathbf{F}_{UV}^{(t)} \right),
\setlength\belowdisplayskip{0pt}
\end{equation}
where $\mathbf{F}_{UV}^{(t)}$ denotes the UV features extracted from frame $t$, and $\textbf{MLP}_{\phi}$ is a lightweight fusion module. The UV tokens are then updated with image features and the new UV features $\mathbf{F}_{UV}^{'(t)}$ following the same process as in Sec.~\ref{encoderdecoder}. The updated tokens are fused with the output tokens of the current frame using the following equation:
\setlength\abovedisplayskip{1pt}
\setlength\belowdisplayskip{0pt}
\begin{align}
\mathbf{T}_{UV}^{'(t)} =& (\mathbf{C}^{'(t-1)} \land \lnot \mathbf{C}^{(t)})\odot\mathbf{T}_{UV}^{'(t-1)} + ( \lnot\mathbf{C}^{'(t-1)} \land \mathbf{C}^{(t)}) \odot \mathbf{T}_{UV}^t \notag\\
+ &( \mathbf{C}^{'(t-1)} \land \mathbf{C}^{(t)}) \odot \textbf{MLP}_{\psi}(\mathbf{T}_{UV}^{'(t-1)}\oplus \mathbf{T}_{UV}^t).
\end{align}
Here, $\mathbf{C}^{(t)}$ measures the visibility of each token in the current view. Visible regions $\mathbf{C}^{'(t-1)}$ are updated by incorporating visible areas from the current frame into the historical visibility map. This prevents degradation caused by partial observations and allows the representation to improve progressively as new frames arrive. The fused tokens are then decoded into a 3D Gaussian representation for the current frame as $\mathcal{G}_{3D}^{(t)} = \mathbf{D}_{3DGS}(\mathbf{T}_{UV}^{'(t)})$. In this way, our method remains practical for both sparse multi-view fusion and long streaming sequences.

\section{Experiment}
\label{exp}

\subsection{Experiment Details}
\label{exp_details}

\textbf{Dataset}. Our method, \textbf{\methodname}, is trained on the VFHQ dataset~\citep{vfhq}, the Nersemble-v2 dataset~\citep{nersemble}, the Ava256 dataset~\citep{ava256}, and our self-curated MV-VFHQ dataset. To construct the MV-VFHQ dataset, we employ the Qwen-Image-Edit~\citep{qwenimageedit} image generator to produce a per-frame multi-view image set, comprising four distinct novel views for each input image. The detailed procedure for generating and annotating the MV-VFHQ dataset is provided in the Appendix.

\label{training}
\noindent\textbf{Training}. We conduct all of the training processes on $8\times$H20 GPUs. The training of \textbf{\methodname} is decomposed into three stages. First, we train a static reconstruction variant of \textbf{\methodname} under a per-frame static setting using the MV-VFHQ, Nersemble-v2, and Ava256 datasets. Second, we adapt this pipeline into an autoregressive reconstruction framework by further fine-tuning under a multi-view setting, again on the MV-VFHQ, Nersemble-v2, and Ava256 datasets. Finally, we fine-tune the model to enable smooth animation, supervised with sequential images from the VFHQ, Nersemble-v2, and Ava256 datasets.
The model is trained with a composite loss function comprising MSE loss, LPIPS loss, mask loss, and attention regularization loss, formulated as:
\begin{equation}
\setlength\abovedisplayskip{1pt}
\mathcal{L}_{all}=\textbf{MSE}(\bm{I}_{render}, \bm{\hat{I}}_{gt})+\textbf{LPIPS}(\bm{I}_{render}, \bm{\hat{I}}_{gt})+\textbf{MSE}(\bm{M}_{render}, \bm{\hat{M}}_{gt})+\mathcal{L}_{attn}.
\setlength\belowdisplayskip{-1pt}
\end{equation}

\noindent\textbf{Metrics}. To evaluate rendered novel-view image quality, we adopt the most commonly used image metrics: PSNR, SSIM, and LPIPS. For identity similarity (CSIM), we compute the cosine distance of face recognition features following \cite{8953658}. To assess expression and pose fidelity, we employ the Average Expression Distance (AED) and Average Pose Distance (APD) as estimated by a 3D Morphable Model (3DMM) \cite{gagavatar}. Additionally, we measure the Average Keypoint Distance (AKD) using a facial landmark detector \cite{8237378}. These metrics provide insights into the accuracy of driving control in our animations. For cross-identity reenactment, where ground truth images are unavailable, we rely on CSIM, AED, and APD for evaluation. These metrics align with those used in prior work \cite{lam}, enabling consistent comparisons across different approaches.

\begin{table}[!htb]
    \centering
        \vspace{-3mm}
    \caption{\textbf{Overview of experimental results}. We evaluate \textbf{\methodname} on 4 different tasks and 3 different datasets. $\bm{e}$, $\bm{v}$, and $\bm{i}$ represents novel \phantom{}expression, view and identity respectively.}
        \vspace{-2mm}
    \resizebox{0.85\linewidth}{!}{\begin{tabular}{lrlrl}
    \hline
    Task& \#Inputs&Evaluation Data&Evaluation Type& Tabs.\& Figs. \\\hline
    Single View Avatar Creation (Self-reenactment)     & 1 &VHFQ & $\bm{e}$&Tab.\ref{tab:vfhq}, Fig.\ref{fig:vfhq}\\
    \phantom{----}- Novel View Assessment     & 1 &Nersemble-v2&$\bm{e}$, $\bm{v}$&Tab.\ref{tab:nersemble}, Fig.\ref{fig:nersmava256}\\
    Single View Avatar Creation (Cross-reenactment)     & 1 &VHFQ&$\bm{e}$, $\bm{i}$&Tab.\ref{tab:vfhq}, Fig.\ref{fig:vfhq}\\
    Multi-View Avatar Creation (Self-reenactment)     & 4 &Ava256&$\bm{e}$, $\bm{v}$&Tab.\ref{tab:ava256}, Fig.\ref{fig:nersmava256}\\
    4D Human Head Reconstruction & stream &Nersemble-v2& $\bm{v}$&Tab.\ref{tab:recon}\\ \hline
    \end{tabular}}
    \vspace{-4mm}
    \label{tab:conclusion}
\end{table}

\subsection{Main Results}
\textbf{Overview}. We carry out experiments on different settings with different datasets to evaluate the performance of our model under various situations. We list our experimental settings in Tab.\ref{tab:conclusion}.

\noindent\textbf{Baselines}. We ensure fair comparisons with all state-of-the-art, open-source head avatar reconstruction models that are directly comparable. For the monocular setting, we compare against LAM~\citep{lam}, GAGAvatar~\citep{gagavatar}, Portrait4D-v2~\citep{portrait4dv2}, Real3DPortrait~\citep{real3d}, and GPAvatar~\citep{gpavatar}. In the multi-view setting, we select InvertAvatar~\citep{invertavatar}, GPAvatar~\citep{gpavatar} DiffusionRig~\citep{diffusionrig}, and FastAvatar~\citep{fastavatar} as baselines. For 4D reconstruction, we include Forge4D~\citep{forge4d}, PanoLAM~\citep{panolam}, and LAM~\citep{lam}. Notably, Forge4D requires four input views to achieve good reconstruction, whereas PanoLAM and LAM take only a single RGB image as input, consistent with our approach.

\begin{figure}[!htb]
    \centering
    \vspace{-4mm}
    \includegraphics[width=0.9\linewidth]{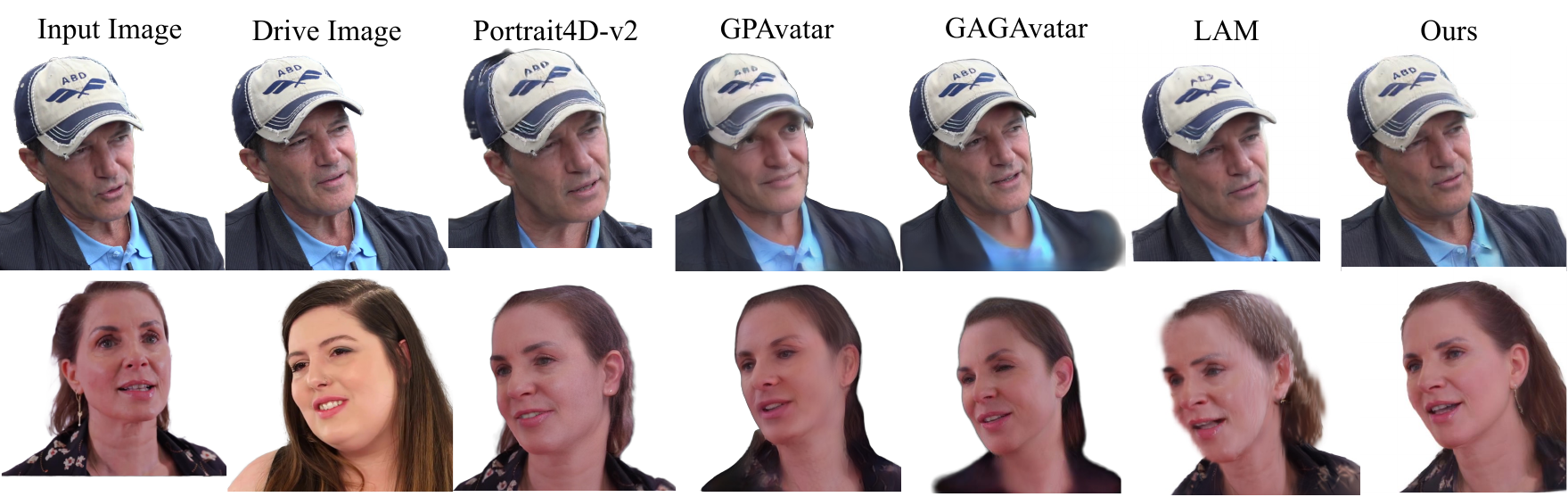}
    \vspace{-3mm}
    \caption{\textbf{Qualitative visualizations on the VFHQ dataset.}}
    \label{fig:vfhq}
    \vspace{-7mm}
\end{figure}

\noindent\textbf{Single-Image Avatar Creation.} We evaluate on VFHQ for self and cross-reenactment (Tab.~\ref{tab:vfhq}, Fig.~\ref{fig:vfhq}). Our model matches prior methods on self-reenactment and surpasses all on cross-reenactment, with maximum gains of +0.27 dB (1.28\%) in PSNR and 0.029 (2.37\%) in CSIM. On NeRSemble-v2 (Tab.~\ref{tab:nersemble}), our model achieves best novel-view quality for novel views. Inference time is 0.4644s on an H20 GPU, comparable to feed-forward methods and faster than optimization-based approaches.

\begin{table}[!htb]
    \centering
    \caption{Comparisons for single-image avatar creation on the VFHQ dataset. This experiment focuses on evaluating models' ability in self-renactment and cross-renactment.}
    \vspace{-2mm}
    \resizebox{0.9\linewidth}{!}{\begin{tabular}{ccccccccccc}
    \hline
            & \multicolumn{7}{c}{Self-Renactment} & \multicolumn{3}{c}{Cross-Renactment} \\ \cline{2-11} 
     & PSNR$\uparrow$   & SSIM$\uparrow$   & LPIPS$\downarrow$  &CSIM$\uparrow$ & AED$\downarrow$ & APD$\downarrow$ &AKD$\downarrow$ &CSIM$\uparrow$ & AED$\downarrow $& APD$\downarrow$     \\ \hline
    GPAvatar   &21.04 &0.807& 0.150 &0.772& 0.132 &0.189& 4.226& 0.564& 0.255& 0.328\\
    Real3DPortrait    &20.88 &0.780 &0.154 &0.801 &0.150 &0.268 &5.971&\underline{0.663} &0.296 &0.411 \\
    Portrait4D-v2    &21.34& 0.791& 0.144& 0.803& 0.117& 0.187& 3.749& 0.656& 0.268& 0.273\\
    GAGAvatar    &21.83& 0.818& 0.122& 0.816 &\underline{0.111} &0.135&3.349& 0.633& 0.253& \underline{0.247}   \\
    LAM    & \underline{22.65}    & \underline{0.829}   &  \underline{0.109}  &   \underline{0.822}   & \textbf{0.102}  &\underline{0.134}& \textbf{2.059}& 0.651& \underline{0.250}&  0.356    \\ \hline
    Ours    &\textbf{22.91}   & \textbf{0.857}  & \textbf{0.096}   &\textbf{0.844}  &0.117&\textbf{0.112}&\underline{2.464}&   \textbf{0.692} &   \textbf{0.229}& \textbf{0.234}  \\ \hline
    \end{tabular}}
    \vspace{-5mm}
    \label{tab:vfhq}
\end{table}

\begin{table}[!htb]
    \centering
    \caption{Comparisons on self-renactment for single-image avatar creation on the Nersemble-v2 dataset. This experiment focuses on evaluating models' ability in novel view synthesis. }
    \vspace{-2mm}
    \resizebox{0.7\linewidth}{!}{
    \begin{tabular}{cccccccc}
    \hline
     & PSNR$\uparrow$   & SSIM$\uparrow$   & LPIPS$\downarrow$  &CSIM$\uparrow$ & AED$\downarrow$ & APD$\downarrow$ &AKD$\downarrow$    \\ \hline
    GPAvatar &20.23 &0.888  & 0.302  & 0.522 & 0.541   &0.535&14.745   \\
    Real3DPortrait & 19.35  &  0.882  & 0.355   & 0.512   &  0.554       &0.554& 13.343     \\
    Portrait4D-v2 & 19.23   & 0.871  &0.330 &  0.538& \underline{0.662}    & 0.592    &36.579         \\
    GAGAvatar & \underline{23.38}  & \underline{0.909}  & 0.164  & \underline{0.688}   & 0.454 &\underline{0.207}&\underline{8.644}          \\
    LAM    & 18.89  &  0.787  & \underline{0.142 } &  0.532    & 0.332  &0.347& 33.214         \\ \hline
    Ours    & \textbf{25.80} & \textbf{0.949} & \textbf{0.091}       &\textbf{0.758} &\textbf{0.102}& \textbf{0.125}& \textbf{6.714}   \\ \hline
    \end{tabular}
    }
    \vspace{-4mm}
    \label{tab:nersemble}
\end{table}

\begin{table}[!htb]
    \centering
    \caption{Comparisons for single- and few-image avatar creation on the Ava256 dataset. This experiment focuses on evaluating models' flexibility in changing the input settings. FFD, GAN, and DIF mean the method is based on feed-forward network, GAN inversion, or Diffusion, respectively.}
    \vspace{-2mm}
    \resizebox{0.8\linewidth}{!}{
    \begin{tabular}{ccccccccc}
    \hline
    \textbf{One-Shot}&Type& PSNR$\uparrow$   & SSIM$\uparrow$   & LPIPS$\downarrow$  &CSIM$\uparrow$ & AED$\downarrow$ & APD$\downarrow$ &AKD$\downarrow$     \\ \hline
    Portrait4D-v2   &FFD& 18.02 & \underline{0.751} & \underline{0.174} & \textbf{0.614} & 0.163 & \underline{0.373} & 14.894         \\
    GAGAvatar    &FFD& 20.37 & 0.702 & 0.237 & 0.525 & \textbf{0.151} & 0.407 & 18,585         \\
    LAM    &FFD& 19.00 & 0.640 & 0.271 & 0.477 & 0.181 & 0.527 & 15.491         \\
    InvertAvatar    &GAN& 20.20 & 0.714 & 0.213 & 0.412 & 0.180 & 0.466 & 12.124     \\
    GPAvatar    &FFD& 20.37 & 0.716 & 0.260 & 0.402 & 0.190 & 0.500 & 23.409         \\
    DiffusionRig   &DIF & \underline{21.39} & 0.738 & 0.251 & 0.403 & 0.211 & 0.423 & \underline{8.876 }         \\ 
    FastAvatar    &FFD& 17.43 & 0.555 & 0.328 & 0.393 & 0.192 & 0.535 & 19.549          \\ 
    Ours    &FFD& \textbf{22.59}  & \textbf{0.750}  & \textbf{0.148} &\underline{0.530} &\underline{0.162}&\textbf{0.360}  &\textbf{5.834}   \\ \hline
     \textbf{Few-Shot} &Type& PSNR$\uparrow$   & SSIM$\uparrow$   & LPIPS$\downarrow$  &CSIM$\uparrow$ & AED$\downarrow$ & APD$\downarrow$ &AKD$\downarrow$    \\ \hline
    InvertAvatar    &GAN&  21.32 &  0.744 &  \underline{0.199} & 0.457 & \underline{0.172} & 0.526 & 11.807          \\
    GPAvatar    &FFD& 19.47 & 0.695 & 0.274 & 0.363 & 0.191 & 0.482 & 21.096          \\
    DiffusionRig    &DIF& \underline{22.72} & \underline{0.777} & 0.235 & \underline{0.503} & 0.186 & \underline{0.380} & \underline{9.859}          \\ 
    FastAvatar    &FFD& 17.51 & 0.548 & 0.308 & 0.461 & 0.195 & 0.562 & 19.557          \\ 
    Ours    &FFD& \textbf{23.01}  & \textbf{0.788}  & \textbf{0.112}   & \textbf{0.673}    &\textbf{0.143}  &\textbf{0.298}&\textbf{3.982 }   \\ \hline
    \end{tabular}
    }
    \vspace{-4mm}
    \label{tab:ava256}
\end{table}

\begin{table}[!htb]
    \centering
    \caption{Comparisons for 4D streaming reconstruction on the Nersemble-v2 dataset. This experiment demonstrates \textbf{\methodname}'s ability in 4D full-head reconstruction.}
    \vspace{-1mm}
    \resizebox{0.7\linewidth}{!}{
    \begin{tabular}{cccccc}
    \hline 
     &Temporal Align&Input Views& PSNR$\uparrow$   & SSIM$\uparrow$   & LPIPS$\downarrow$     \\ \hline
    Forge4D & Yes&4& 19.32 & 0.712 & 0.146 \\
    PanoLAM & No&1& 20.17  & 0.917  &0.214  \\
    LAM    & No&1&19.96   & 0.799   &0.145   \\ \hline
    Ours    & Yes&1&25.33&0.942&0.096    \\ \hline
    \end{tabular}
    }
    \vspace{-4mm}
    \label{tab:recon}
\end{table}

\begin{figure}[t]
    \vspace{-3mm}
    \centering
     \includegraphics[width=0.9\linewidth]{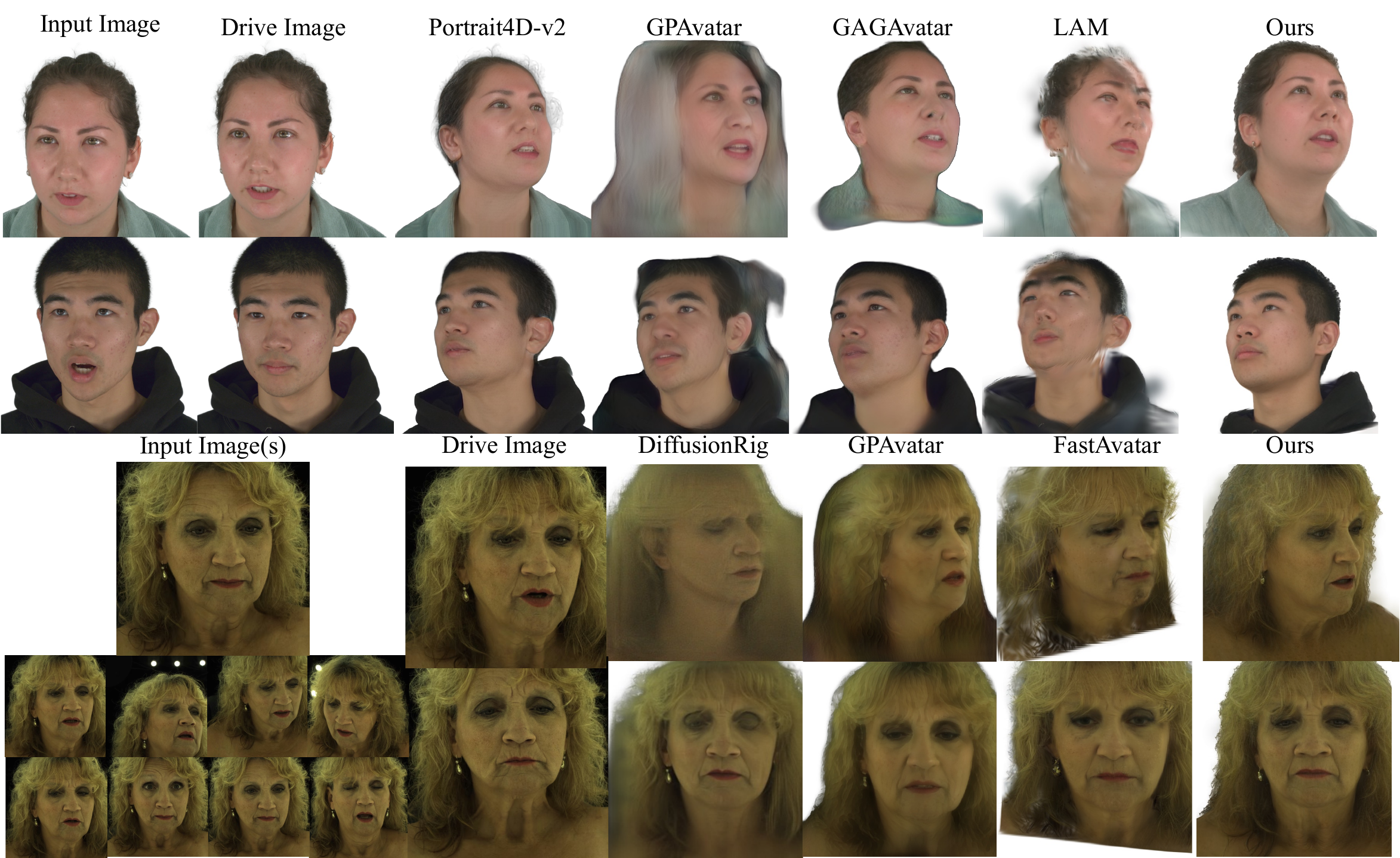}
    \vspace{-3mm}
    \caption{\textbf{Qualitative visualizations on Nersemble-v2 and Ava256 datasets.}}
    \vspace{-5mm}
    \label{fig:nersmava256}
\end{figure}

\noindent\textbf{Few-Image Avatar Creation}.
We evaluate few-shot avatar creation performance on the Ava256 dataset. Quantitative comparisons are summarized in Tab.~\ref{tab:ava256}, and qualitative results are illustrated in Fig.~\ref{fig:nersmava256}. Our model consistently outperforms previous methods across all metrics. Additionally, creating an avatar from few-shot images yields a noticeable performance gain of +0.42 dB in PSNR.

\noindent\textbf{Static Full-Head Reconstruction}. We demonstrate the static per-frame full-head reconstruction capability of \textbf{\methodname} by comparing novel view image qualities of reconstructed human heads on the NeRSemble-v2 test set. We show metrical reports in Tab.~\ref{tab:recon}. Our method outperforms other methods in reconstructed static human head quality.

\subsection{Ablation Study}
\begin{figure}[!htb]
    \centering
    \vspace{-4mm}
    \includegraphics[width=0.84\linewidth]{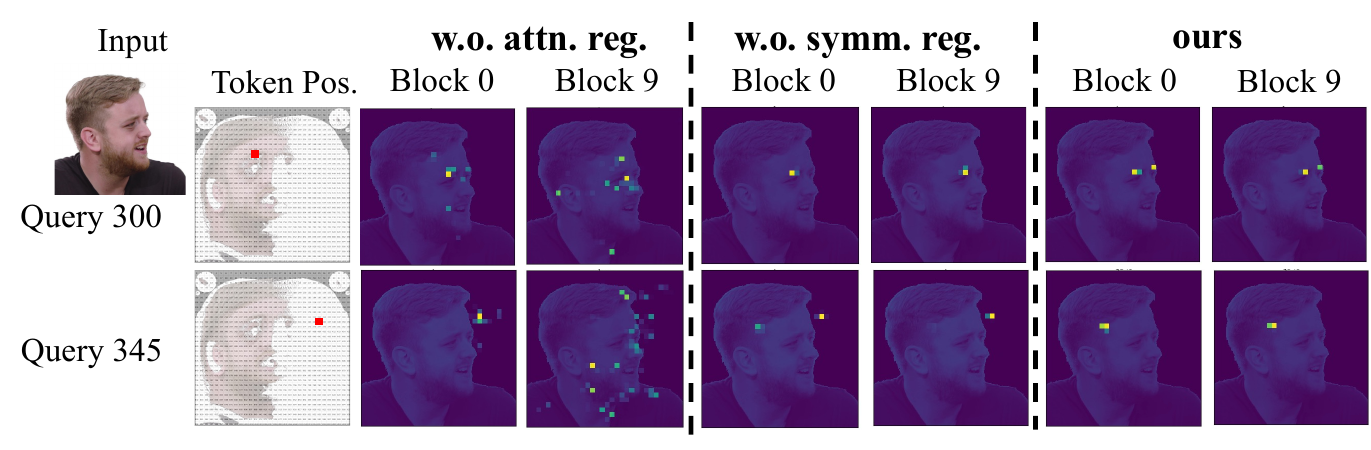}
    \vspace{-5.5mm}
    \caption{Visualizations on the attention activation maps of different attention regularization methods.
    }
    \vspace{-3mm}
    \label{fig:abl_attnvis}
\end{figure}

\begin{figure}[H]
    \centering
    \includegraphics[width=0.9\linewidth]{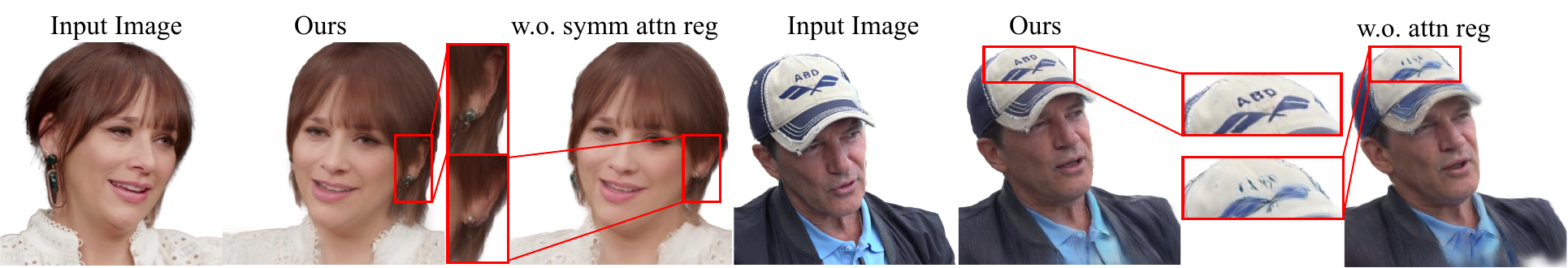}
    \vspace{-3.6mm}
    \caption{Qualitative visualization on different attention regularization supervisions. }
    \vspace{-3.5mm}
    \label{fig:abl_attn_reg}
\end{figure}

\noindent\textbf{Effectiveness of Attention Regularization}. We evaluate our attention regularization by comparing four variants: 1) no regularization; 2) semantic regularization only; 3) full regularization; and 4) direct DINO feature warping (replacing cross-attention). Attention activation visualizations (Fig.~\ref{fig:abl_attnvis}) show that our method achieves better activation with correct focus. Quantitative results (Tab.~\ref{tab:abl_attnreg}) and qualitative comparisons (Fig.~\ref{fig:abl_attn_reg}) demonstrate that facial details are significantly improved, with bilateral symmetry issues persisting only under semantic regularization alone.

\begin{figure}[t]
    \centering
    \includegraphics[width=0.85\linewidth]{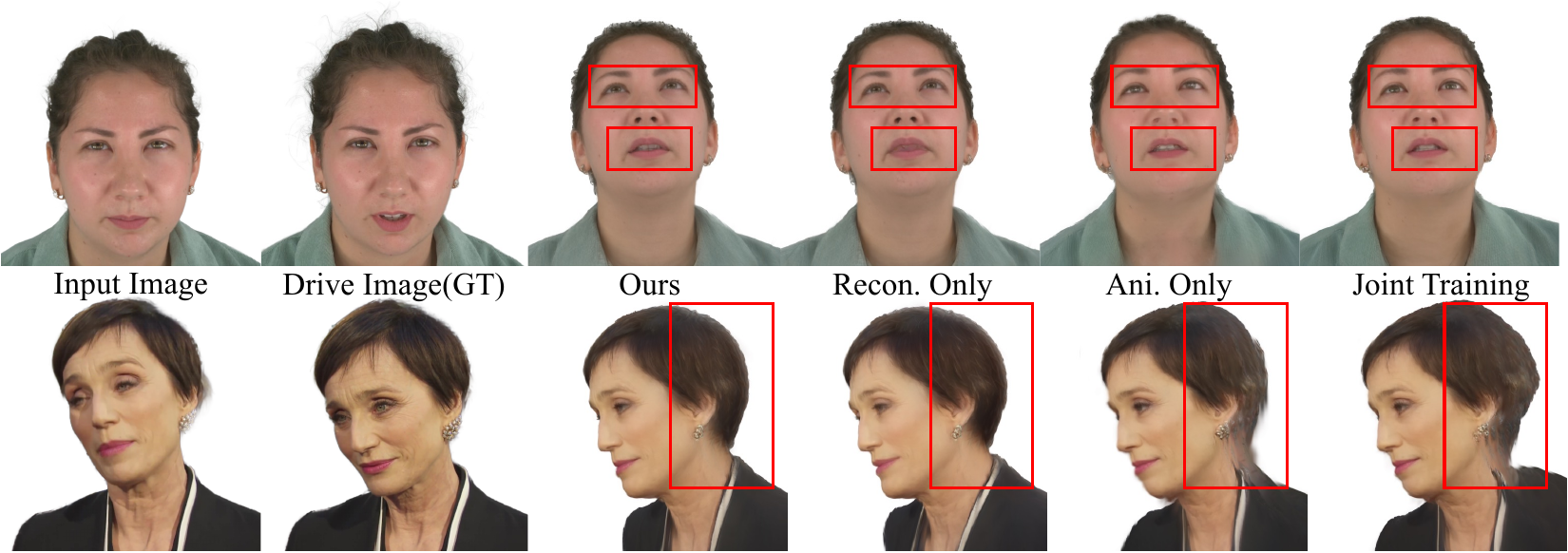}
    \vspace{-3.5mm}
    \caption{Qualitative visualization of different training strategies.}
    \vspace{-4.2mm}
    \label{fig:abl_2stages}
\end{figure}

\noindent\textbf{Effectiveness of the Dual-Phase Training Strategy}. We demonstrate the effectiveness of our dual-phase training strategy by comparing the following training variants: 1) joint supervision of novel views and novel expressions; 2) training with novel expressions only; 3) training with novel views only; and 4) separating training into a reconstruction stage followed by an animation stage. Quantitative results and qualitative comparisons are presented in Tab.~\ref{tab:abl_reconanim} and Fig.~\ref{fig:abl_2stages}, respectively. The results indicate that our proposed attention regularization achieves better symmetrical hallucination, while the full semantic attention regularization leads to improved recovery of fine facial details.
\label{rta}

\begin{table}[!htb]
    \centering
    \begin{minipage}{0.46\linewidth}
        \centering
        \caption{Ablation studies on different training strategies.}
        \vspace{-2.5mm}
        \resizebox{\linewidth}{!}{
        \begin{tabular}{lccc}
            \hline
            & PSNR$\uparrow$ & SSIM$\uparrow$ & LPIPS$\downarrow$ \\ \hline
            Joint training           & 24.99 & 0.938 & 0.103 \\
            Animation Only           & 24.93 & 0.940 & 0.111 \\
            Reconstruction Only      & 25.65 & 0.939 & 0.099 \\ \hline
            Recon-then-Ani (Ours) & 25.80 & 0.949 & 0.091 \\ \hline
        \end{tabular}
        }
        \label{tab:abl_reconanim}
    \end{minipage}
    \hfill
    \begin{minipage}{0.46\linewidth}
        \centering
        \vspace{-3mm}
        \caption{Ablation studies on different attention regularization strategies.}
        \vspace{-2mm}
        \resizebox{\linewidth}{!}{
        \begin{tabular}{lccc}
            \hline
            & PSNR$\uparrow$ & SSIM$\uparrow$ & LPIPS$\downarrow$ \\ \hline
            direct feature unwarping         & 23.34 & 0.871 & 0.157 \\ 
            w.o. attn-reg           & 25.30 & 0.934 & 0.101 \\
            w. sym. attn-reg        & 25.68 & 0.932 & 0.096 \\ \hline
            w. full attn-reg (Ours) & 25.80 & 0.949 & 0.091 \\ \hline
        \end{tabular}
        }
        \vspace{-4mm}
        \label{tab:abl_attnreg}
    \end{minipage}
    \vspace{-3mm}
\end{table}

\begin{table}[H]
    \centering
    \caption{Quantitative results of 4D head reconstruction on Nersemble-v2 dataset with streaming inputs. We use TL-STD and TI to measure the temporal performance.}
    \vspace{-2.5mm}
    \resizebox{\linewidth}{!}{
    \begin{tabular}{cccccccc}
    \hline
     & TL-STD$\downarrow$& TI$\downarrow$& PSNR$\uparrow$   & SSIM$\uparrow$   & LPIPS$\downarrow$  & GPU Memory$\downarrow$     \\ \hline
    Full-Attention    &-&-&-&-&-&OOM\\
    Framewise Reconstruction        &8.141&13.731&24.46&0.940&0.106&8751MB\\
    Auto-regressive (No Token Fusion) &8.253&13.760&24.53&0.940&0.099&8755MB\\\hline 
    Auto-regressive (Visibility-Aware Fusion)  &7.253&12.805&25.33&0.942&0.096&8771MB\\\hline 
    \end{tabular}
    }
    \vspace{-4mm}
    \label{tab:abl_ar}
\end{table}

\begin{figure}[H]
    \centering
    \includegraphics[width=0.8\linewidth]{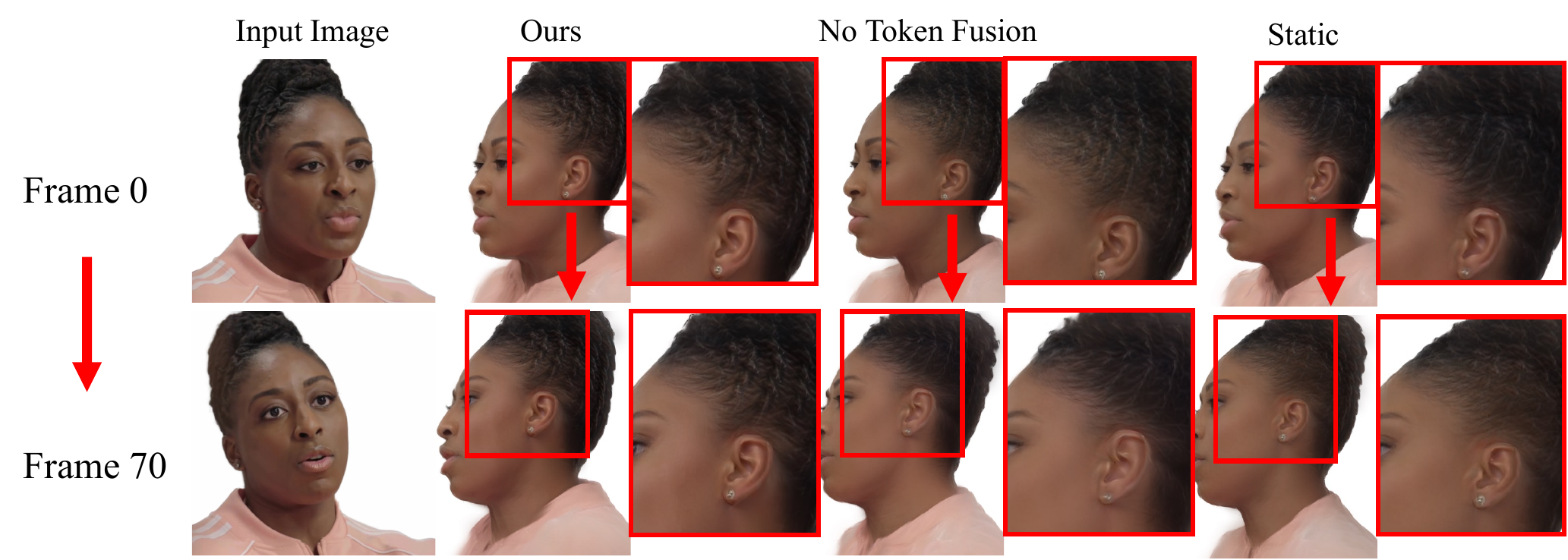}
    \vspace{-4mm}
    \caption{Qualitative evaluation on our auto-regressive token fusion designed. Our design enables model to generate temporal consistent hallucinations in invisible areas.}
    \label{abl:ar}
    \vspace{-5mm}
\end{figure}

\noindent\textbf{Efficiency of Autoregressive Adaptation}. We evaluate our autoregressive adaptation by comparing animated novel-view image quality and GPU memory consumption in Tab.~\ref{tab:abl_ar}. Directly reconstructing 4D heads from a large number of streaming frames (e.g., 20 frames) causes out-of-memory issues. In contrast, our autoregressive fusion model reconstructs 4D heads with minimal and constant additional GPU cost (+20 MB, 0.228\%) and achieves more temporally stable results, as evidenced by gains in Temporal Luminance Standard Deviation (TL-STD)~\citep{arend1976response} (+0.888) and Temporal Information (TI, ITU-T P.910~\citep{itu-t_p.910_handle}) (+0.926).
\section{Conclusion}
\label{conclusion}
In this paper, we propose a 3D Gaussian-based focus-aware large full-head avatar reconstruction model that places greater emphasis on the semantic and symmetric properties of the human head. In addition, we present a comprehensive analysis of the degradation in large-view reconstruction performance when the model is supervised simultaneously with both novel-view and novel-time images, identifying this as an entanglement between the tasks of reconstruction and animation. Furthermore, we extend the proposed pipeline to support streaming inputs through an efficient autoregressive formulation. Future works may focus on improving the quality at occiputs and extending this pipeline to full-body avatars.

\medskip

\bibliographystyle{unsrtnat}
\bibliography{main}

\begin{thebibliography}{66}
\providecommand{\natexlab}[1]{#1}
\providecommand{\url}[1]{\texttt{#1}}
\expandafter\ifx\csname urlstyle\endcsname\relax
  \providecommand{\doi}[1]{doi: #1}\else
  \providecommand{\doi}{doi: \begingroup \urlstyle{rm}\Url}\fi

\bibitem[Qian et~al.(2024)Qian, Kirschstein, Schoneveld, Davoli, Giebenhain, and Nie{\ss}ner]{gaussianavatars}
Shenhan Qian, Tobias Kirschstein, Liam Schoneveld, Davide Davoli, Simon Giebenhain, and Matthias Nie{\ss}ner.
\newblock Gaussianavatars: Photorealistic head avatars with rigged 3d gaussians.
\newblock In \emph{Proceedings of the IEEE/CVF Conference on Computer Vision and Pattern Recognition}, pages 20299--20309, 2024.

\bibitem[Deng et~al.(2024{\natexlab{a}})Deng, Wang, Ren, Chen, and Wang]{portrait4d}
Yu~Deng, Duomin Wang, Xiaohang Ren, Xingyu Chen, and Baoyuan Wang.
\newblock Portrait4d: Learning one-shot 4d head avatar synthesis using synthetic data.
\newblock In \emph{IEEE/CVF Conference on Computer Vision and Pattern Recognition}, 2024{\natexlab{a}}.

\bibitem[Deng et~al.(2024{\natexlab{b}})Deng, Wang, and Wang]{portrait4dv2}
Yu~Deng, Duomin Wang, and Baoyuan Wang.
\newblock Portrait4d-v2: Pseudo multi-view data creates better 4d head synthesizer.
\newblock \emph{arXiv preprint arXiv:2403.13570}, 2024{\natexlab{b}}.

\bibitem[Zhang et~al.(2025{\natexlab{a}})Zhang, Wu, Liang, Gong, Hu, Yao, Cao, and Zhu]{fate}
Jiawei Zhang, Zijian Wu, Zhiyang Liang, Yicheng Gong, Dongfang Hu, Yao Yao, Xun Cao, and Hao Zhu.
\newblock Fate: Full-head gaussian avatar with textural editing from monocular video.
\newblock In \emph{Proceedings of the Computer Vision and Pattern Recognition Conference}, pages 5535--5545, 2025{\natexlab{a}}.

\bibitem[Lee et~al.(2025)Lee, Kang, Buehler, Kim, Hwang, Hyung, Jang, and Choo]{surfhead}
Jaeseong Lee, Taewoong Kang, Marcel Buehler, Min-Jung Kim, Sungwon Hwang, Junha Hyung, Hyojin Jang, and Jaegul Choo.
\newblock Surfhead: Affine rig blending for geometrically accurate 2d gaussian surfel head avatars.
\newblock In \emph{The Thirteenth International Conference on Learning Representations}, 2025.
\newblock URL \url{https://openreview.net/forum?id=1x1gGg49jr}.

\bibitem[Xu et~al.(2023)Xu, Wang, Zhao, Zhang, and Liu]{avatarmav}
Yuelang Xu, Lizhen Wang, Xiaochen Zhao, Hongwen Zhang, and Yebin Liu.
\newblock Avatarmav: Fast 3d head avatar reconstruction using motion-aware neural voxels.
\newblock In \emph{ACM SIGGRAPH 2023 Conference Proceedings}, 2023.

\bibitem[Xu et~al.(2024{\natexlab{a}})Xu, Wang, Zheng, Su, and Liu]{xu20243d}
Yuelang Xu, Lizhen Wang, Zerong Zheng, Zhaoqi Su, and Yebin Liu.
\newblock 3d gaussian parametric head model.
\newblock In \emph{European Conference on Computer Vision}, pages 129--147. Springer, 2024{\natexlab{a}}.

\bibitem[Zheng et~al.(2025)Zheng, Wen, Li, Zhang, Su, Chang, Zhao, Lv, Zhang, Zhang, Wang, and Xu]{headgap}
Xiaozheng Zheng, Chao Wen, Zhaohu Li, Weiyi Zhang, Zhuo Su, Xu~Chang, Yang Zhao, Zheng Lv, Xiaoyuan Zhang, Yongjie Zhang, Guidong Wang, and Lan Xu.
\newblock { HeadGAP: Few-Shot 3D Head Avatar via Generalizable Gaussian Priors }.
\newblock In \emph{2025 International Conference on 3D Vision (3DV)}, pages 946--957, Los Alamitos, CA, USA, March 2025. IEEE Computer Society.
\newblock \doi{10.1109/3DV66043.2025.00092}.
\newblock URL \url{https://doi.ieeecomputersociety.org/10.1109/3DV66043.2025.00092}.

\bibitem[Yu et~al.(2024)Yu, Bai, Meka, Tan, Xu, Pandey, Fanello, Park, and Zhang]{one2avatar}
Zhixuan Yu, Ziqian Bai, Abhimitra Meka, Feitong Tan, Qiangeng Xu, Rohit Pandey, Sean Fanello, Hyun~Soo Park, and Yinda Zhang.
\newblock One2avatar: Generative implicit head avatar for few-shot user adaptation, 2024.
\newblock URL \url{https://arxiv.org/abs/2402.11909}.

\bibitem[Chen et~al.(2024)Chen, Wang, Li, Xiao, Zhang, Yao, and Liu]{chen2024monogaussianavatar}
Yufan Chen, Lizhen Wang, Qijing Li, Hongjiang Xiao, Shengping Zhang, Hongxun Yao, and Yebin Liu.
\newblock Monogaussianavatar: Monocular gaussian point-based head avatar.
\newblock In \emph{ACM SIGGRAPH 2024 conference papers}, pages 1--9, 2024.

\bibitem[Hong et~al.(2024)Hong, Zhang, Gu, Bi, Zhou, Liu, Liu, Sunkavalli, Bui, and Tan]{lrm}
Yicong Hong, Kai Zhang, Jiuxiang Gu, Sai Bi, Yang Zhou, Difan Liu, Feng Liu, Kalyan Sunkavalli, Trung Bui, and Hao Tan.
\newblock {LRM}: Large reconstruction model for single image to 3d.
\newblock In \emph{The Twelfth International Conference on Learning Representations}, 2024.
\newblock URL \url{https://openreview.net/forum?id=sllU8vvsFF}.

\bibitem[He et~al.(2025)He, Gu, Ye, Xu, Zhao, Dong, Yuan, Dong, and Bo]{lam}
Yisheng He, Xiaodong Gu, Xiaodan Ye, Chao Xu, Zhengyi Zhao, Yuan Dong, Weihao Yuan, Zilong Dong, and Liefeng Bo.
\newblock Lam: Large avatar model for one-shot animatable gaussian head.
\newblock In \emph{Proceedings of the Special Interest Group on Computer Graphics and Interactive Techniques Conference Conference Papers}, pages 1--13, 2025.

\bibitem[He and Hoi(2026)]{he2026meshlam}
Yisheng He and Steven Hoi.
\newblock Meshlam: Feed-forward one-shot animatable textured mesh avatar reconstruction, 2026.
\newblock URL \url{https://arxiv.org/abs/2604.22865}.

\bibitem[Li et~al.(2025)Li, He, Hu, Dong, Yuan, Liu, Zhu, Cheng, Dong, and Guo]{panolam}
Peng Li, Yisheng He, Yingdong Hu, Yuan Dong, Weihao Yuan, Yuan Liu, Siyu Zhu, Gang Cheng, Zilong Dong, and Yike Guo.
\newblock Panolam: Large avatar model for gaussian full-head synthesis from one-shot unposed image, 2025.
\newblock URL \url{https://arxiv.org/abs/2509.07552}.

\bibitem[Kirschstein et~al.(2025{\natexlab{a}})Kirschstein, Romero, Sevastopolsky, Nie{\ss}ner, and Saito]{avat3r}
Tobias Kirschstein, Javier Romero, Artem Sevastopolsky, Matthias Nie{\ss}ner, and Shunsuke Saito.
\newblock Avat3r: Large animatable gaussian reconstruction model for high-fidelity 3d head avatars.
\newblock In \emph{Proceedings of the IEEE/CVF International Conference on Computer Vision (ICCV)}, pages 12089--12100, October 2025{\natexlab{a}}.

\bibitem[Li et~al.(2017)Li, Bolkart, Black, Li, and Romero]{flame}
Tianye Li, Timo Bolkart, Michael.~J. Black, Hao Li, and Javier Romero.
\newblock Learning a model of facial shape and expression from {4D} scans.
\newblock \emph{ACM Transactions on Graphics, (Proc. SIGGRAPH Asia)}, 36\penalty0 (6):\penalty0 194:1--194:17, 2017.
\newblock URL \url{https://doi.org/10.1145/3130800.3130813}.

\bibitem[Hu et~al.(2025)Hu, He, Chen, Yuan, Qiu, Lin, Zhu, Dong, and Zhang]{forge4d}
Yingdong Hu, Yisheng He, Jinnan Chen, Weihao Yuan, Kejie Qiu, Zehong Lin, Siyu Zhu, Zilong Dong, and Jun Zhang.
\newblock Forge4d: Feed-forward 4d human reconstruction and interpolation from uncalibrated sparse-view videos, 2025.
\newblock URL \url{https://arxiv.org/abs/2509.24209}.

\bibitem[Wang et~al.(2025)Wang, Chen, Karaev, Vedaldi, Rupprecht, and Novotny]{vggt}
Jianyuan Wang, Minghao Chen, Nikita Karaev, Andrea Vedaldi, Christian Rupprecht, and David Novotny.
\newblock Vggt: Visual geometry grounded transformer.
\newblock In \emph{Proceedings of the IEEE/CVF Conference on Computer Vision and Pattern Recognition}, 2025.

\bibitem[Zhuo et~al.(2025)Zhuo, Zheng, Guo, Wu, Zhou, and Lu]{streamvggt}
Dong Zhuo, Wenzhao Zheng, Jiahe Guo, Yuqi Wu, Jie Zhou, and Jiwen Lu.
\newblock Streaming 4d visual geometry transformer.
\newblock \emph{arXiv preprint arXiv:2507.11539}, 2025.

\bibitem[Xie et~al.(2022)Xie, Wang, Zhang, Dong, and Shan]{vfhq}
Liangbin Xie, Xintao Wang, Honglun Zhang, Chao Dong, and Ying Shan.
\newblock Vfhq: A high-quality dataset and benchmark for video face super-resolution.
\newblock In \emph{The IEEE Conference on Computer Vision and Pattern Recognition Workshops (CVPRW)}, 2022.

\bibitem[Wu et~al.(2025)Wu, Li, Zhou, Lin, Gao, Yan, ming Yin, Bai, Xu, Chen, Chen, Tang, Zhang, Wang, Yang, Yu, Cheng, Liu, Li, Zhang, Meng, Wei, Ni, Chen, Cao, Peng, Qu, Wu, Wang, Yu, Wen, Feng, Xu, Wang, Zhang, Zhu, Wu, Cai, and Liu]{qwenimageedit}
Chenfei Wu, Jiahao Li, Jingren Zhou, Junyang Lin, Kaiyuan Gao, Kun Yan, Sheng ming Yin, Shuai Bai, Xiao Xu, Yilei Chen, Yuxiang Chen, Zecheng Tang, Zekai Zhang, Zhengyi Wang, An~Yang, Bowen Yu, Chen Cheng, Dayiheng Liu, Deqing Li, Hang Zhang, Hao Meng, Hu~Wei, Jingyuan Ni, Kai Chen, Kuan Cao, Liang Peng, Lin Qu, Minggang Wu, Peng Wang, Shuting Yu, Tingkun Wen, Wensen Feng, Xiaoxiao Xu, Yi~Wang, Yichang Zhang, Yongqiang Zhu, Yujia Wu, Yuxuan Cai, and Zenan Liu.
\newblock Qwen-image technical report, 2025.
\newblock URL \url{https://arxiv.org/abs/2508.02324}.

\bibitem[Goodfellow et~al.(2014)Goodfellow, Pouget-Abadie, Mirza, Xu, Warde-Farley, Ozair, Courville, and Bengio]{goodfellow2014generative}
Ian~J Goodfellow, Jean Pouget-Abadie, Mehdi Mirza, Bing Xu, David Warde-Farley, Sherjil Ozair, Aaron Courville, and Yoshua Bengio.
\newblock Generative adversarial nets.
\newblock \emph{Advances in neural information processing systems}, 27, 2014.

\bibitem[Burkov et~al.(2020)Burkov, Pasechnik, Grigorev, and Lempitsky]{burkov2020neural}
Egor Burkov, Igor Pasechnik, Artur Grigorev, and Victor Lempitsky.
\newblock Neural head reenactment with latent pose descriptors.
\newblock In \emph{Proceedings of the IEEE/CVF conference on computer vision and pattern recognition}, pages 13786--13795, 2020.

\bibitem[Wang et~al.(2023)Wang, Deng, Yin, Shum, and Wang]{wang2023progressive}
Duomin Wang, Yu~Deng, Zixin Yin, Heung-Yeung Shum, and Baoyuan Wang.
\newblock Progressive disentangled representation learning for fine-grained controllable talking head synthesis.
\newblock In \emph{Proceedings of the IEEE/CVF Conference on Computer Vision and Pattern Recognition}, pages 17979--17989, 2023.

\bibitem[Zakharov et~al.(2019)Zakharov, Shysheya, Burkov, and Lempitsky]{zakharov2019few}
Egor Zakharov, Aliaksandra Shysheya, Egor Burkov, and Victor Lempitsky.
\newblock Few-shot adversarial learning of realistic neural talking head models.
\newblock In \emph{Proceedings of the IEEE/CVF international conference on computer vision}, pages 9459--9468, 2019.

\bibitem[Hong et~al.(2022)Hong, Zhang, Shen, and Xu]{hong2022depth}
Fa-Ting Hong, Longhao Zhang, Li~Shen, and Dan Xu.
\newblock Depth-aware generative adversarial network for talking head video generation.
\newblock In \emph{Proceedings of the IEEE/CVF conference on computer vision and pattern recognition}, pages 3397--3406, 2022.

\bibitem[Karras et~al.(2019)Karras, Laine, and Aila]{karras2019style}
Tero Karras, Samuli Laine, and Timo Aila.
\newblock A style-based generator architecture for generative adversarial networks.
\newblock In \emph{Proceedings of the IEEE/CVF conference on computer vision and pattern recognition}, pages 4401--4410, 2019.

\bibitem[Drobyshev et~al.(2022)Drobyshev, Chelishev, Khakhulin, Ivakhnenko, Lempitsky, and Zakharov]{drobyshev2022megaportraits}
Nikita Drobyshev, Jenya Chelishev, Taras Khakhulin, Aleksei Ivakhnenko, Victor Lempitsky, and Egor Zakharov.
\newblock Megaportraits: One-shot megapixel neural head avatars.
\newblock In \emph{Proceedings of the 30th ACM International Conference on Multimedia}, pages 2663--2671, 2022.

\bibitem[Guo et~al.(2024)Guo, Zhang, Liu, Zhong, Zhang, Wan, and Zhang]{guo2024liveportrait}
Jianzhu Guo, Dingyun Zhang, Xiaoqiang Liu, Zhizhou Zhong, Yuan Zhang, Pengfei Wan, and Di~Zhang.
\newblock Liveportrait: Efficient portrait animation with stitching and retargeting control.
\newblock \emph{arXiv preprint arXiv:2407.03168}, 2024.

\bibitem[Zhang et~al.(2023)Zhang, Qi, Zhang, Zhang, Wu, Chen, Chen, Wang, and Wen]{zhang2023metaportrait}
Bowen Zhang, Chenyang Qi, Pan Zhang, Bo~Zhang, HsiangTao Wu, Dong Chen, Qifeng Chen, Yong Wang, and Fang Wen.
\newblock Metaportrait: Identity-preserving talking head generation with fast personalized adaptation.
\newblock In \emph{Proceedings of the IEEE/CVF Conference on Computer Vision and Pattern Recognition}, pages 22096--22105, 2023.

\bibitem[Tian et~al.(2024)Tian, Wang, Zhang, and Bo]{tian2024emo}
Linrui Tian, Qi~Wang, Bang Zhang, and Liefeng Bo.
\newblock Emo: Emote portrait alive generating expressive portrait videos with audio2video diffusion model under weak conditions.
\newblock In \emph{European Conference on Computer Vision}, pages 244--260. Springer, 2024.

\bibitem[Cui et~al.(2024)Cui, Li, Yao, Zhu, Shang, Cheng, Zhou, Zhu, and Wang]{cui2024hallo2}
Jiahao Cui, Hui Li, Yao Yao, Hao Zhu, Hanlin Shang, Kaihui Cheng, Hang Zhou, Siyu Zhu, and Jingdong Wang.
\newblock Hallo2: Long-duration and high-resolution audio-driven portrait image animation.
\newblock \emph{arXiv preprint arXiv:2410.07718}, 2024.

\bibitem[Xu et~al.(2024{\natexlab{b}})Xu, Li, Su, Shang, Zhang, Liu, Wang, Yao, and Zhu]{xu2024hallo}
Mingwang Xu, Hui Li, Qingkun Su, Hanlin Shang, Liwei Zhang, Ce~Liu, Jingdong Wang, Yao Yao, and Siyu Zhu.
\newblock Hallo: Hierarchical audio-driven visual synthesis for portrait image animation.
\newblock \emph{arXiv preprint arXiv:2406.08801}, 2024{\natexlab{b}}.

\bibitem[Taubner et~al.(2025{\natexlab{a}})Taubner, Zhang, Tuli, and Lindell]{cap4d}
Felix Taubner, Ruihang Zhang, Mathieu Tuli, and David~B. Lindell.
\newblock {CAP4D}: Creating animatable {4D} portrait avatars with morphable multi-view diffusion models.
\newblock In \emph{Proceedings of the IEEE/CVF Conference on Computer Vision and Pattern Recognition (CVPR)}, pages 5318--5330, June 2025{\natexlab{a}}.

\bibitem[Taubner et~al.(2025{\natexlab{b}})Taubner, Zhang, Tuli, Bahmani, and Lindell]{mvp4d}
Felix Taubner, Ruihang Zhang, Mathieu Tuli, Sherwin Bahmani, and David~B Lindell.
\newblock Mvp4d: Multi-view portrait video diffusion for animatable 4d avatars.
\newblock In \emph{Proceedings of the SIGGRAPH Asia 2025 Conference Papers}, pages 1--11, 2025{\natexlab{b}}.

\bibitem[Loper et~al.(2023)Loper, Mahmood, Romero, Pons-Moll, and Black]{loper2023smpl}
Matthew Loper, Naureen Mahmood, Javier Romero, Gerard Pons-Moll, and Michael~J Black.
\newblock Smpl: A skinned multi-person linear model.
\newblock In \emph{Seminal Graphics Papers: Pushing the Boundaries, Volume 2}, pages 851--866. 2023.

\bibitem[Pavlakos et~al.(2019)Pavlakos, Choutas, Ghorbani, Bolkart, Osman, Tzionas, and Black]{SMPL-X:2019}
Georgios Pavlakos, Vasileios Choutas, Nima Ghorbani, Timo Bolkart, Ahmed A.~A. Osman, Dimitrios Tzionas, and Michael~J. Black.
\newblock Expressive body capture: 3d hands, face, and body from a single image.
\newblock In \emph{Proceedings IEEE Conf. on Computer Vision and Pattern Recognition (CVPR)}, 2019.

\bibitem[Mildenhall et~al.(2021)Mildenhall, Srinivasan, Tancik, Barron, Ramamoorthi, and Ng]{nerf}
Ben Mildenhall, Pratul~P Srinivasan, Matthew Tancik, Jonathan~T Barron, Ravi Ramamoorthi, and Ren Ng.
\newblock Nerf: Representing scenes as neural radiance fields for view synthesis.
\newblock \emph{Communications of the ACM}, 65\penalty0 (1):\penalty0 99--106, 2021.

\bibitem[Chu et~al.(2024{\natexlab{a}})Chu, Li, Zeng, Yang, Lin, Liu, and Harada]{chu2024gpavatar}
Xuangeng Chu, Yu~Li, Ailing Zeng, Tianyu Yang, Lijian Lin, Yunfei Liu, and Tatsuya Harada.
\newblock {GPA}vatar: Generalizable and precise head avatar from image(s).
\newblock In \emph{The Twelfth International Conference on Learning Representations}, 2024{\natexlab{a}}.
\newblock URL \url{https://openreview.net/forum?id=hgehGq2bDv}.

\bibitem[Kirschstein et~al.(2024)Kirschstein, Giebenhain, Tang, Georgopoulos, and Nie\ss{}ner]{kirschstein2024gghead}
Tobias Kirschstein, Simon Giebenhain, Jiapeng Tang, Markos Georgopoulos, and Matthias Nie\ss{}ner.
\newblock {GGHead: Fast and Generalizable 3D Gaussian Heads}.
\newblock In \emph{SIGGRAPH Asia 2024 Conference Papers}, SA '24, New York, NY, USA, 2024. Association for Computing Machinery.
\newblock ISBN 9798400711312.
\newblock \doi{10.1145/3680528.3687686}.
\newblock URL \url{https://doi.org/10.1145/3680528.3687686}.

\bibitem[Li et~al.(2023)Li, Zhang, Wang, Zhao, Wang, Chen, Zhang, Wang, Bo, and Li]{10203662}
Weichuang Li, Longhao Zhang, Dong Wang, Bin Zhao, Zhigang Wang, Mulin Chen, Bang Zhang, Zhongjian Wang, Liefeng Bo, and Xuelong Li.
\newblock { One-Shot High-Fidelity Talking-Head Synthesis with Deformable Neural Radiance Field }.
\newblock In \emph{2023 IEEE/CVF Conference on Computer Vision and Pattern Recognition (CVPR)}, pages 17969--17978, Los Alamitos, CA, USA, June 2023. IEEE Computer Society.
\newblock \doi{10.1109/CVPR52729.2023.01723}.
\newblock URL \url{https://doi.ieeecomputersociety.org/10.1109/CVPR52729.2023.01723}.

\bibitem[Kerbl et~al.(2023)Kerbl, Kopanas, Leimk{\"u}hler, Drettakis, et~al.]{3dgs}
Bernhard Kerbl, Georgios Kopanas, Thomas Leimk{\"u}hler, George Drettakis, et~al.
\newblock 3d gaussian splatting for real-time radiance field rendering.
\newblock \emph{ACM Trans. Graph.}, 42\penalty0 (4):\penalty0 139--1, 2023.

\bibitem[Xu et~al.(2024{\natexlab{c}})Xu, Chen, Li, Zhang, Wang, Zheng, and Liu]{gaussianheadavatar}
Yuelang Xu, Benwang Chen, Zhe Li, Hongwen Zhang, Lizhen Wang, Zerong Zheng, and Yebin Liu.
\newblock Gaussian head avatar: Ultra high-fidelity head avatar via dynamic gaussians.
\newblock In \emph{Proceedings of the IEEE/CVF Conference on Computer Vision and Pattern Recognition (CVPR)}, 2024{\natexlab{c}}.

\bibitem[Liao et~al.(2025)Liao, Zheng, Karmanov, Hu, Jin, Xiu, and Li]{soap}
Tingting Liao, Yujian Zheng, Adilbek Karmanov, Liwen Hu, Leyang Jin, Yuliang Xiu, and Hao Li.
\newblock Soap: Style-omniscient animatable portraits.
\newblock 2025.
\newblock URL \url{https://arxiv.org/abs/2505.05022}.

\bibitem[Chu and Harada(2024)]{gagavatar}
Xuangeng Chu and Tatsuya Harada.
\newblock Generalizable and animatable gaussian head avatar.
\newblock In \emph{The Thirty-eighth Annual Conference on Neural Information Processing Systems}, 2024.
\newblock URL \url{https://openreview.net/forum?id=gVM2AZ5xA6}.

\bibitem[Zheng et~al.(2024)Zheng, Zhou, Shao, Liu, Zhang, Nie, and Liu]{gpsgs}
Shunyuan Zheng, Boyao Zhou, Ruizhi Shao, Boning Liu, Shengping Zhang, Liqiang Nie, and Yebin Liu.
\newblock Gps-gaussian: Generalizable pixel-wise 3d gaussian splatting for real-time human novel view synthesis.
\newblock In \emph{Proceedings of the IEEE/CVF Conference on Computer Vision and Pattern Recognition (CVPR)}, 2024.

\bibitem[Hu et~al.(2024)Hu, Liu, Shao, Lin, and Zhang]{evags}
Yingdong Hu, Zhening Liu, Jiawei Shao, Zehong Lin, and Jun Zhang.
\newblock Eva-gaussian: 3d gaussian-based real-time human novel view synthesis under diverse camera settings, 2024.
\newblock URL \url{https://arxiv.org/abs/2410.01425}.

\bibitem[Peng et~al.(2025)Peng, Su, Wang, Guo, Li, Long, Lv, Sun, Zhang, and Liu]{flexavatarpeng}
Cheng Peng, Zhuo Su, Liao Wang, Chen Guo, Zhaohu Li, Chengjiang Long, Zheng Lv, Jingxiang Sun, Chenyangguang Zhang, and Yebin Liu.
\newblock Flexavatar: Flexible large reconstruction model for animatable gaussian head avatars with detailed deformation, 2025.
\newblock URL \url{https://arxiv.org/abs/2512.17717}.

\bibitem[Kirschstein et~al.(2025{\natexlab{b}})Kirschstein, Giebenhain, and Nie{\ss}ner]{flexavatartobias}
Tobias Kirschstein, Simon Giebenhain, and Matthias Nie{\ss}ner.
\newblock Flexavatar: Learning complete 3d head avatars with partial supervision.
\newblock \emph{arXiv preprint arXiv:2512.15599}, 2025{\natexlab{b}}.

\bibitem[Wang et~al.(2024)Wang, Leroy, Cabon, Chidlovskii, and Revaud]{dust3r}
Shuzhe Wang, Vincent Leroy, Yohann Cabon, Boris Chidlovskii, and Jerome Revaud.
\newblock Dust3r: Geometric 3d vision made easy.
\newblock In \emph{CVPR}, 2024.

\bibitem[Wu et~al.(2026{\natexlab{a}})Wu, Zhou, Hu, Liu, Sun, Wang, Cao, Shen, and Zhu]{uika}
Zijian Wu, Boyao Zhou, Liangxiao Hu, Hongyu Liu, Yuan Sun, Xuan Wang, Xun Cao, Yujun Shen, and Hao Zhu.
\newblock Uika: Fast universal head avatar from pose-free images, 2026{\natexlab{a}}.
\newblock URL \url{https://arxiv.org/abs/2601.07603}.

\bibitem[Zhang et~al.(2025{\natexlab{b}})Zhang, Chu, Li, Zang, Li, Li, Cao, Zhu, and Lu]{bringingportrait}
Jiawei Zhang, Lei Chu, Jiahao Li, Zhenyu Zang, Chong Li, Xiao Li, Xun Cao, Hao Zhu, and Yan Lu.
\newblock Bringing your portrait to 3d presence, 2025{\natexlab{b}}.
\newblock URL \url{https://arxiv.org/abs/2511.22553}.

\bibitem[Yu et~al.(2020)Yu, Kumar, Gupta, Levine, Hausman, and Finn]{pcgrad}
Tianhe Yu, Saurabh Kumar, Abhishek Gupta, Sergey Levine, Karol Hausman, and Chelsea Finn.
\newblock Gradient surgery for multi-task learning.
\newblock \emph{arXiv preprint arXiv:2001.06782}, 2020.

\bibitem[van~der Maaten and Hinton(2008)]{tsne}
Laurens van~der Maaten and Geoffrey Hinton.
\newblock Visualizing data using t-sne.
\newblock \emph{Journal of Machine Learning Research}, 9\penalty0 (86):\penalty0 2579--2605, 2008.
\newblock URL \url{http://jmlr.org/papers/v9/vandermaaten08a.html}.

\bibitem[Kirschstein et~al.(2023)Kirschstein, Qian, Giebenhain, Walter, and Nie\ss{}ner]{nersemble}
Tobias Kirschstein, Shenhan Qian, Simon Giebenhain, Tim Walter, and Matthias Nie\ss{}ner.
\newblock Nersemble: Multi-view radiance field reconstruction of human heads.
\newblock \emph{ACM Trans. Graph.}, 42\penalty0 (4), jul 2023.
\newblock ISSN 0730-0301.
\newblock \doi{10.1145/3592455}.
\newblock URL \url{https://doi.org/10.1145/3592455}.

\bibitem[Martinez et~al.(2024)Martinez, Kim, Romero, Bagautdinov, Saito, Yu, Anderson, Zollhöfer, Wang, Bai, Li, Wei, Joshi, Borsos, Simon, Saragih, Theodosis, Greene, Josyula, Maeta, Jewett, Venshtain, Heilman, Chen, Fu, Elshaer, Du, Wu, Chen, Kang, Wu, Emad, Longay, Brewer, Shah, Booth, Koska, Haidle, Andromalos, Hsu, Dauer, Selednik, Godisart, Ardisson, Cipperly, Humberston, Farr, Hansen, Guo, Braun, Krenn, Wen, Evans, Fadeeva, Stewart, Schwartz, Gupta, Moon, Guo, Dong, Xu, Shiratori, Prada, Pires, Peng, Buffalini, Trimble, McPhail, Schoeller, and Sheikh]{ava256}
Julieta Martinez, Emily Kim, Javier Romero, Timur Bagautdinov, Shunsuke Saito, Shoou-I Yu, Stuart Anderson, Michael Zollhöfer, Te-Li Wang, Shaojie Bai, Chenghui Li, Shih-En Wei, Rohan Joshi, Wyatt Borsos, Tomas Simon, Jason Saragih, Paul Theodosis, Alexander Greene, Anjani Josyula, Silvio~Mano Maeta, Andrew~I. Jewett, Simon Venshtain, Christopher Heilman, Yueh-Tung Chen, Sidi Fu, Mohamed Ezzeldin~A. Elshaer, Tingfang Du, Longhua Wu, Shen-Chi Chen, Kai Kang, Michael Wu, Youssef Emad, Steven Longay, Ashley Brewer, Hitesh Shah, James Booth, Taylor Koska, Kayla Haidle, Matt Andromalos, Joanna Hsu, Thomas Dauer, Peter Selednik, Tim Godisart, Scott Ardisson, Matthew Cipperly, Ben Humberston, Lon Farr, Bob Hansen, Peihong Guo, Dave Braun, Steven Krenn, He~Wen, Lucas Evans, Natalia Fadeeva, Matthew Stewart, Gabriel Schwartz, Divam Gupta, Gyeongsik Moon, Kaiwen Guo, Yuan Dong, Yichen Xu, Takaaki Shiratori, Fabian Prada, Bernardo~R. Pires, Bo~Peng, Julia Buffalini, Autumn Trimble, Kevyn McPhail, Melissa Schoeller, and
  Yaser Sheikh.
\newblock {Codec Avatar Studio: Paired Human Captures for Complete, Driveable, and Generalizable Avatars}.
\newblock \emph{NeurIPS Track on Datasets and Benchmarks}, 2024.

\bibitem[Deng et~al.(2019)Deng, Guo, Xue, and Zafeiriou]{8953658}
Jiankang Deng, Jia Guo, Niannan Xue, and Stefanos Zafeiriou.
\newblock Arcface: Additive angular margin loss for deep face recognition.
\newblock In \emph{2019 IEEE/CVF Conference on Computer Vision and Pattern Recognition (CVPR)}, pages 4685--4694, 2019.
\newblock \doi{10.1109/CVPR.2019.00482}.

\bibitem[Bulat and Tzimiropoulos(2017)]{8237378}
Adrian Bulat and Georgios Tzimiropoulos.
\newblock How far are we from solving the 2d \& 3d face alignment problem? (and a dataset of 230,000 3d facial landmarks).
\newblock In \emph{2017 IEEE International Conference on Computer Vision (ICCV)}, pages 1021--1030, 2017.
\newblock \doi{10.1109/ICCV.2017.116}.

\bibitem[Ye et~al.(2024)Ye, Zhong, Ren, Yang, Li, Huang, Jiang, He, Huang, Liu, et~al.]{real3d}
Zhenhui Ye, Tianyun Zhong, Yi~Ren, Jiaqi Yang, Weichuang Li, Jiawei Huang, Ziyue Jiang, Jinzheng He, Rongjie Huang, Jinglin Liu, et~al.
\newblock Real3d-portrait: One-shot realistic 3d talking portrait synthesis.
\newblock \emph{arXiv preprint arXiv:2401.08503}, 2024.

\bibitem[Chu et~al.(2024{\natexlab{b}})Chu, Li, Zeng, Yang, Lin, Liu, and Harada]{gpavatar}
Xuangeng Chu, Yu~Li, Ailing Zeng, Tianyu Yang, Lijian Lin, Yunfei Liu, and Tatsuya Harada.
\newblock {GPA}vatar: Generalizable and precise head avatar from image(s).
\newblock In \emph{The Twelfth International Conference on Learning Representations}, 2024{\natexlab{b}}.
\newblock URL \url{https://openreview.net/forum?id=hgehGq2bDv}.

\bibitem[Zhao et~al.(2024)Zhao, Sun, Wang, Suo, and Liu]{invertavatar}
Xiaochen Zhao, Jingxiang Sun, Lizhen Wang, Jinli Suo, and Yebin Liu.
\newblock Invertavatar: Incremental gan inversion for generalized head avatars.
\newblock In \emph{ACM SIGGRAPH 2024 Conference Papers}, SIGGRAPH '24, New York, NY, USA, 2024. Association for Computing Machinery.
\newblock ISBN 9798400705250.
\newblock \doi{10.1145/3641519.3657478}.
\newblock URL \url{https://doi.org/10.1145/3641519.3657478}.

\bibitem[Zheng~Ding and Zhang(2023)]{diffusionrig}
Zhihao Xia Lars Jebe Zhuowen~Tu Zheng~Ding, Cecilia~Zhang and Xiuming Zhang.
\newblock Diffusionrig: Learning personalized priors for facial appearance editing.
\newblock In \emph{Proceedings of the IEEE/CVF Conference on Computer Vision and Pattern Recognition}, 2023.

\bibitem[Wu et~al.(2026{\natexlab{b}})Wu, Chen, Wu, Li, Lu, and Feng]{fastavatar}
Yue Wu, Xuanhong Chen, Yufan Wu, Wen Li, Yuxi Lu, and Kairui Feng.
\newblock Fastavatar: Towards unified and fast 3d avatar reconstruction with large gaussian reconstruction transformers.
\newblock In \emph{The Fourteenth International Conference on Learning Representations}, 2026{\natexlab{b}}.
\newblock URL \url{https://openreview.net/forum?id=P7zBSCs4Xt}.

\bibitem[Arend~Jr(1976)]{arend1976response}
Lawrence~E Arend~Jr.
\newblock Response of the human eye to spatially sinusoidal gratings at various exposure durations.
\newblock \emph{Vision Research}, 16\penalty0 (11):\penalty0 1311--1315, 1976.

\bibitem[ITU-T(2022)]{itu-t_p.910_handle}
ITU-T.
\newblock Subjective video quality assessment methods for multimedia applications.
\newblock ITU-T Recommendation P.910, 2022.
\newblock URL \url{https://handle.itu.int/11.1002/1000/15005}.
\newblock Accessed on: 2026-05-05.

\bibitem[Danecek et~al.(2022)Danecek, Black, and Bolkart]{emoca}
Radek Danecek, Michael~J. Black, and Timo Bolkart.
\newblock {EMOCA}: {E}motion driven monocular face capture and animation.
\newblock In \emph{Conference on Computer Vision and Pattern Recognition (CVPR)}, pages 20311--20322, 2022.

\end{thebibliography}

\newpage
\appendix
\section{More Visualizations}

\begin{figure}
    \centering
    \includegraphics[width=\linewidth]{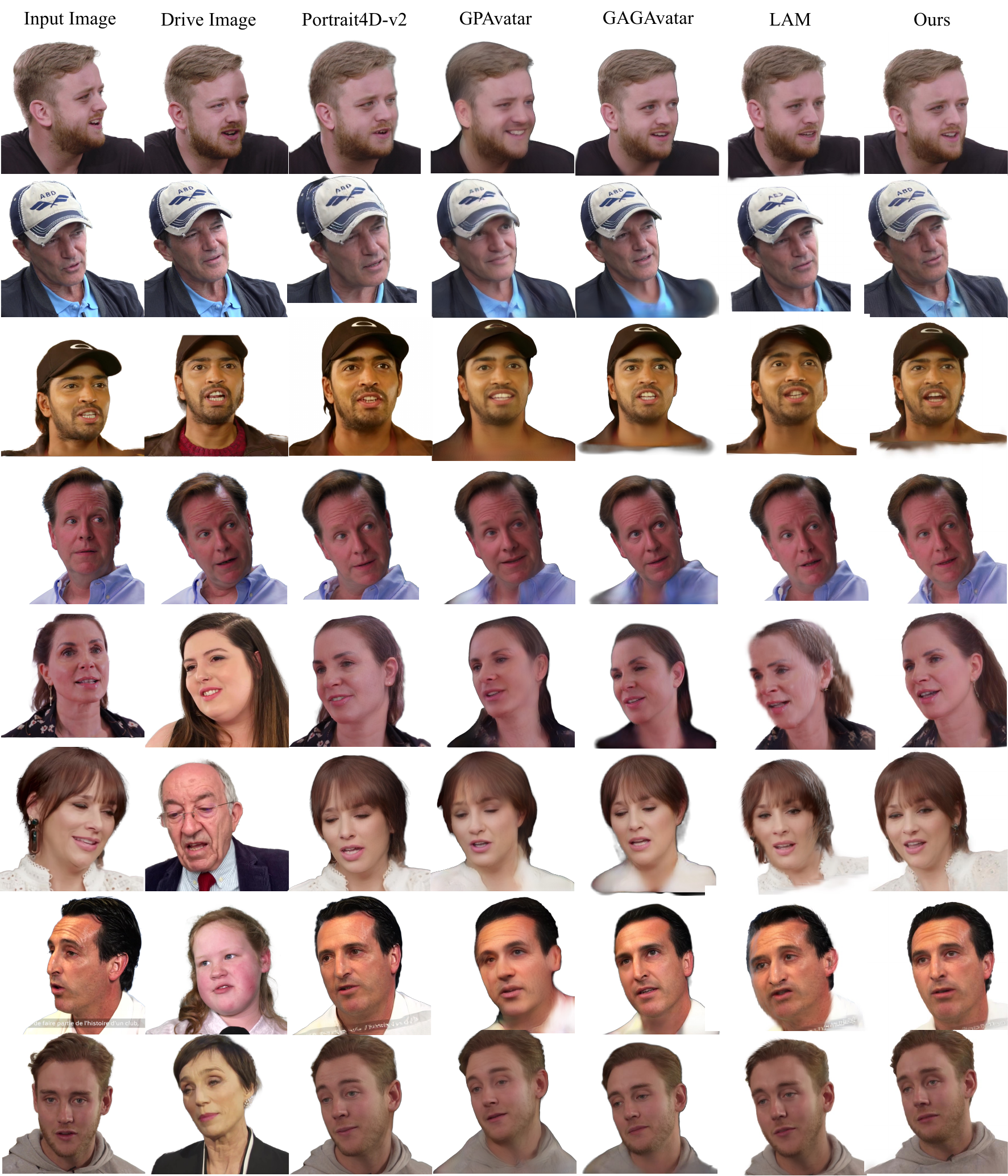}
    \caption{\textbf{Qualitative visualizations on VFHQ dataset.}}
    \label{fig:vfhq_appendix}
\end{figure}

In this section, we provide additional visualizations. Specifically, we show attention maps obtained without attention regularization in Fig.~\ref{fig:attention_no_reg}, attention maps with different attention regularizations in Fig.~\ref{fig:attention_no_reg}, and present more qualitative results for both self-reenactment and cross-reenactment on the VFHQ dataset in Fig.~\ref{fig:vfhq_appendix}.
\begin{figure}[!htb]
    \centering
    \includegraphics[width=0.9\linewidth]{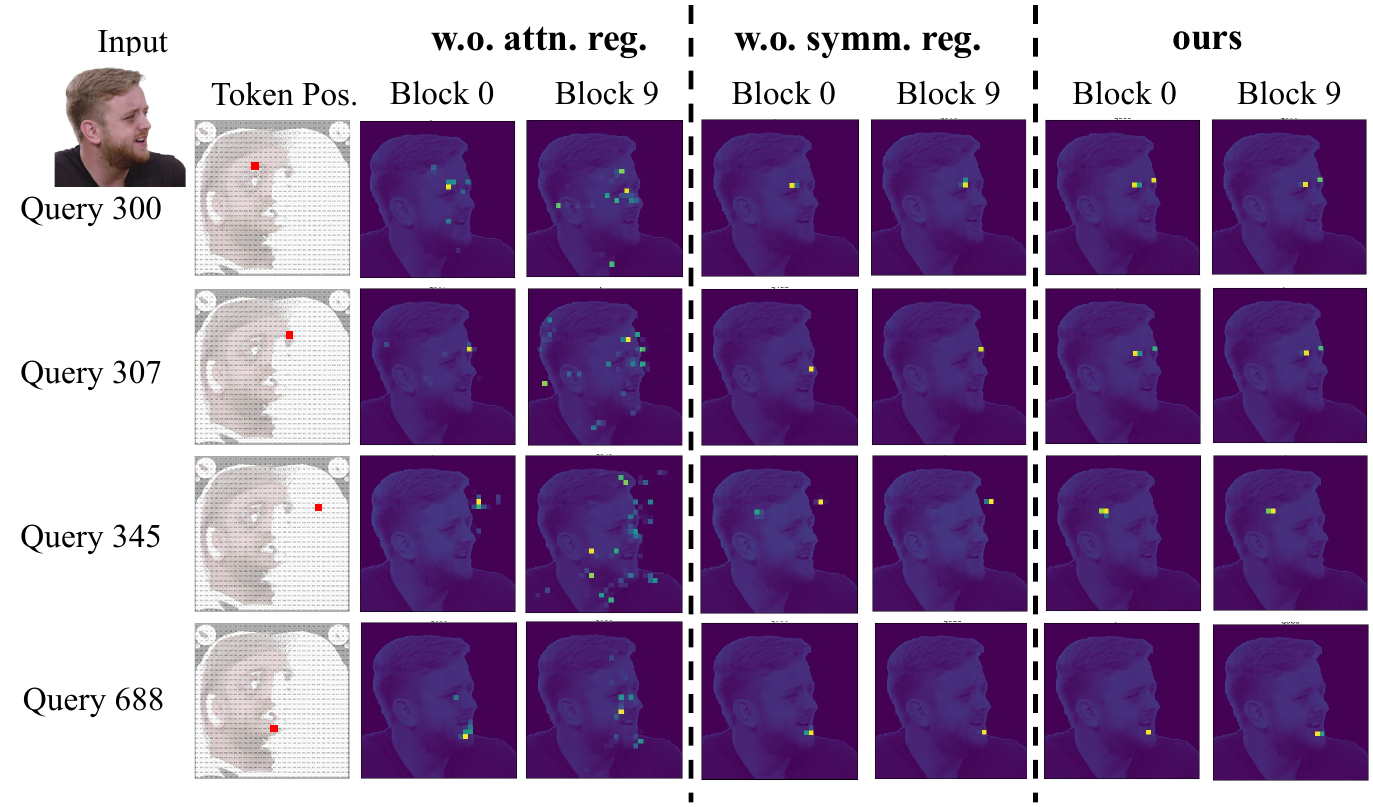}
    \caption{\textbf{Visualization of attention map activations} in a vanilla-trained network. Among the randomly selected tokens, token 147 and token 299 are visible in the input RGB image, whereas token 346, token 547, and token 695 are not. Please zoom in for detailed views and the corresponding token positions on the UV map.}
    \label{fig:attention_no_reg}
\end{figure}

\begin{figure}[!htb]
    \centering
    \includegraphics[width=0.9\linewidth]{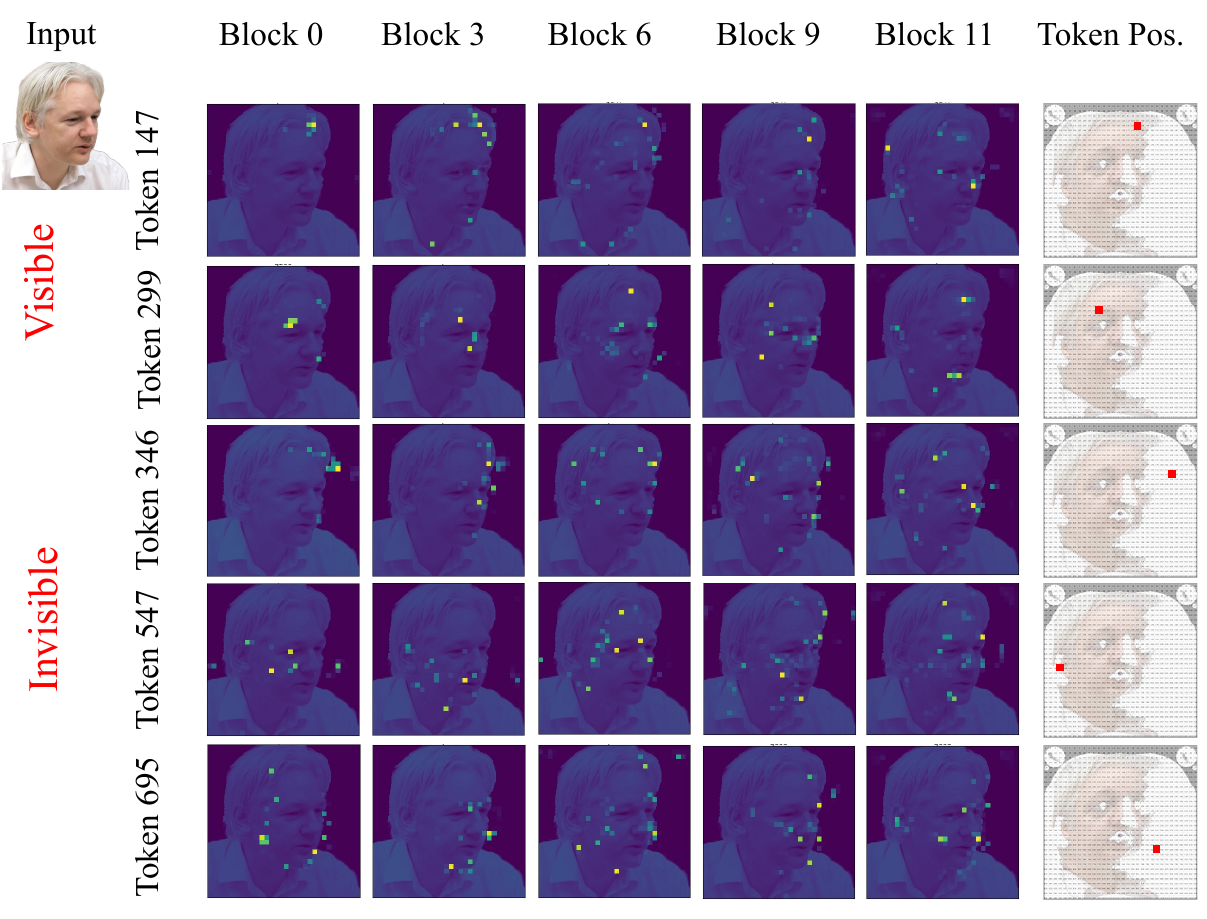}
    \caption{Visualizations on the attention activation maps of different attention regularization methods.}
    \label{fig:attention_no_reg}
\end{figure}

\section{Details on the MV-VFHQ Dataset}

We provide further details on the construction of our MV-VFHQ dataset. For each frame in every video sequence, we feed the image into Qwen-Image-Edit~\citep{qwenimageedit} with four camera control prompts: (1) turn the camera right to $45^\circ$; (2) turn the camera right to $90^\circ$; (3) turn the camera left to $45^\circ$; (4) turn the camera left to $90^\circ$. This process augments the original monocular VFHQ dataset with four additional novel views per frame.

\begin{figure}[!htb]
    \centering
    \includegraphics[width=\linewidth]{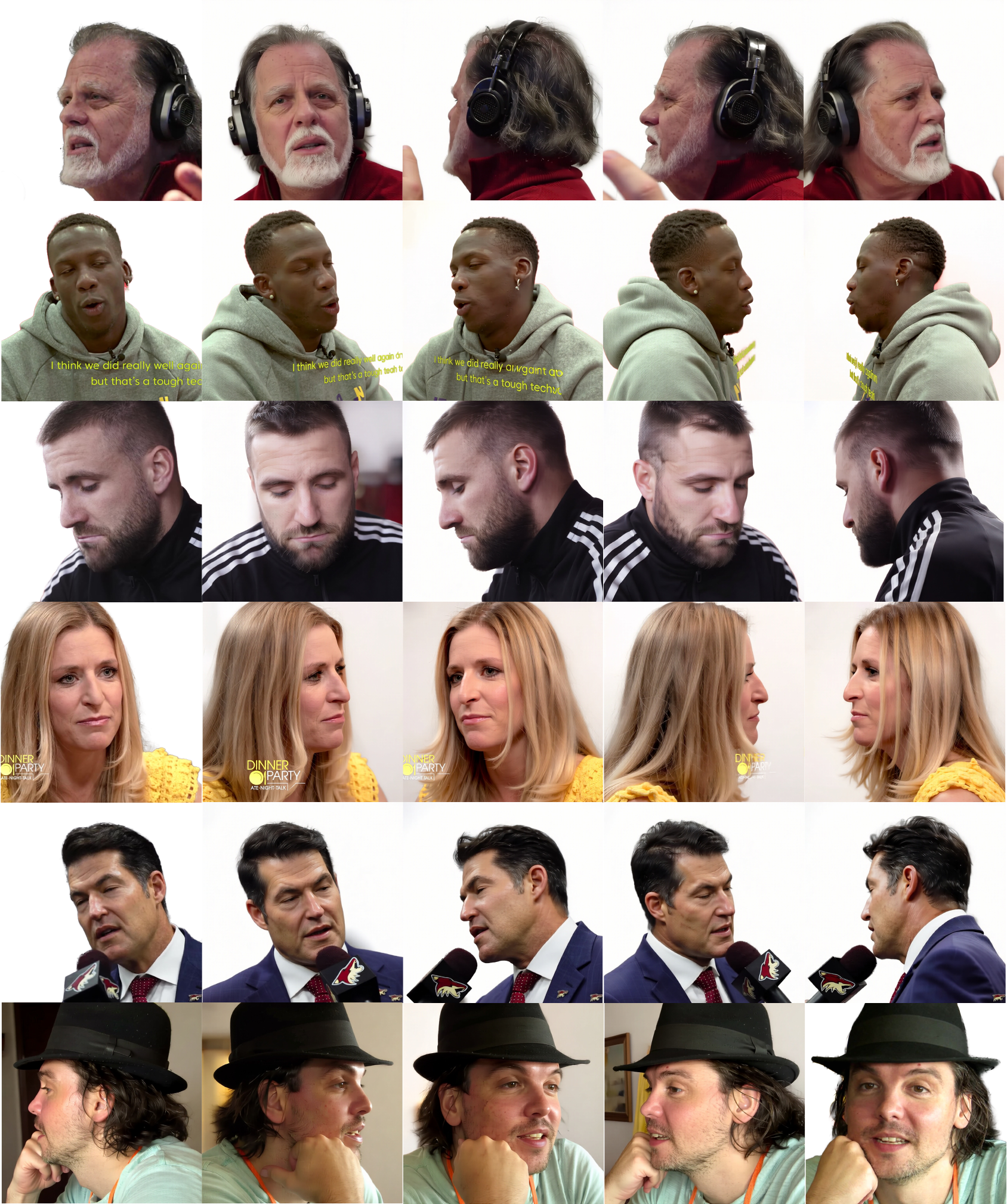}
    \caption{Samples of Multi-view VFHQ dataset.}
    \label{fig:mv_vfhq_sample}
\end{figure}

We additionally annotate each frame with FLAME shape parameters, expression parameters, and camera pose parameters. To obtain these annotations, we employ an optimization-based FLAME tracker that minimizes the distance between labeled human face keypoints and the corresponding FLAME-derived keypoints, with initializations provided by EMOCA~\citep{emoca}. A sample visualization of the resulting dataset is shown in Fig.~\ref{fig:mv_vfhq_sample}.

\section{Experimental Details}
\label{details}
For Nersemble-v2 dataset, we leave all sequences for id 017 and 018 for evaluation, while all others for training. For Ava256 dataset, we leave all sequence for id 20210810--1306--FXN596 for evaluation and all others for training. For VFHQ dataset, we follow the authors' instruction for training and evaluation. We adopt Adam as optimizer, with a initialized learning rate of 2$\times10^{-4}$, which linearly decreases to 1$\times10^{-7}$ at the end of training. 

\section{Broader Impacts}
\label{impact}
\paragraph{Positive Impacts.}
\textbf{\methodname}~democratizes animatable 3D/4D head avatar creation from a single or few images, enabling applications in telepresence, VR/AR, creative industries, and assistive technology. Our analysis of attention mechanisms and task disentanglement also provides insights for future research in generative 3D vision.

\paragraph{Negative Impacts \& Safeguards.}
Our method could be misused for deepfake generation or unauthorized identity reconstruction. To mitigate these risks: (1) all training data are from public datasets with informed consent; (2) code and models are released under a Non-Commercial Academic License prohibiting deceptive or privacy-violating uses.


\newpage
\clearpage
\section*{NeurIPS Paper Checklist}

The checklist is designed to encourage best practices for responsible machine learning research, addressing issues of reproducibility, transparency, research ethics, and societal impact. Do not remove the checklist: {\bf The papers not including the checklist will be desk rejected.} The checklist should follow the references and follow the (optional) supplemental material.  The checklist does NOT count towards the page
limit. 

Please read the checklist guidelines carefully for information on how to answer these questions. For each question in the checklist:
\begin{itemize}
    \item You should answer \answerYes{}, \answerNo{}, or \answerNA{}.
    \item \answerNA{} means either that the question is Not Applicable for that particular paper or the relevant information is Not Available.
    \item Please provide a short (1--2 sentence) justification right after your answer (even for \answerNA). 
\end{itemize}

{\bf The checklist answers are an integral part of your paper submission.} They are visible to the reviewers, area chairs, senior area chairs, and ethics reviewers. You will also be asked to include it (after eventual revisions) with the final version of your paper, and its final version will be published with the paper.

The reviewers of your paper will be asked to use the checklist as one of the factors in their evaluation. While \answerYes{} is generally preferable to \answerNo{}, it is perfectly acceptable to answer \answerNo{} provided a proper justification is given (e.g., error bars are not reported because it would be too computationally expensive'' or ``we were unable to find the license for the dataset we used''). In general, answering \answerNo{} or \answerNA{} is not grounds for rejection. While the questions are phrased in a binary way, we acknowledge that the true answer is often more nuanced, so please just use your best judgment and write a justification to elaborate. All supporting evidence can appear either in the main paper or the supplemental material, provided in appendix. If you answer \answerYes{} to a question, in the justification please point to the section(s) where related material for the question can be found.

IMPORTANT, please:
\begin{itemize}
    \item {\bf Delete this instruction block, but keep the section heading ``NeurIPS Paper Checklist"},
    \item  {\bf Keep the checklist subsection headings, questions/answers and guidelines below.}
    \item {\bf Do not modify the questions and only use the provided macros for your answers}.
\end{itemize}


\begin{enumerate}

\item {\bf Claims}
    \item[] Question: Do the main claims made in the abstract and introduction accurately reflect the paper's contributions and scope?
    \item[] Answer: \answerYes{} 
    \item[] Justification: The abstract and introduction accurately reflect the paper’s contributions by
proposing a symmetrical and semantic attention regularization (Section 3.2), a two stage training method (Section 3.3) and auto-regressive reconstruction (Section 3.4).

\item {\bf Limitations}
    \item[] Question: Does the paper discuss the limitations of the work performed by the authors?
    \item[] Answer: \answerYes{} 
    \item[] Justification: Please see Sec.~\ref{conclusion}.

\item {\bf Theory assumptions and proofs}
    \item[] Question: For each theoretical result, does the paper provide the full set of assumptions and a complete (and correct) proof?
    \item[] Answer: \answerNA{} 
    \item[] Justification: This paper does not cotain any theoretical result.

    \item {\bf Experimental result reproducibility}
    \item[] Question: Does the paper fully disclose all the information needed to reproduce the main experimental results of the paper to the extent that it affects the main claims and/or conclusions of the paper (regardless of whether the code and data are provided or not)?
    \item[] Answer: \answerYes{} 
    \item[] Justification: Experimental details are provided in Sec.~\ref{training}.

\item {\bf Open access to data and code}
    \item[] Question: Does the paper provide open access to the data and code, with sufficient instructions to faithfully reproduce the main experimental results, as described in supplemental material?
    \item[] Answer: \answerNo{} 
    \item[] Justification: This article involves commercial collaboration; the code and data will be released upon review and approval.

\item {\bf Experimental setting/details}
    \item[] Question: Does the paper specify all the training and test details (e.g., data splits, hyperparameters, how they were chosen, type of optimizer) necessary to understand the results?
    \item[] Answer: \answerYes{} 
    \item[] Justification: Please see Appendix Sec.~\ref{details} and Sec.~\ref{exp_details}.

\item {\bf Experiment statistical significance}
    \item[] Question: Does the paper report error bars suitably and correctly defined or other appropriate information about the statistical significance of the experiments?
    \item[] Answer: \answerYes{}
    \item[] Justification: Please see Sec.~\ref{exp}.

\item {\bf Experiments compute resources}
    \item[] Question: For each experiment, does the paper provide sufficient information on the computer resources (type of compute workers, memory, time of execution) needed to reproduce the experiments?
    \item[] Answer: \answerYes{} 
    \item[] Justification: Please see Sec.~\ref{exp}.
    
\item {\bf Code of ethics}
    \item[] Question: Does the research conducted in the paper conform, in every respect, with the NeurIPS Code of Ethics \url{https://neurips.cc/public/EthicsGuidelines}?
    \item[] Answer: \answerYes{} 
    \item[] Justification: We have reviewed and obeyed the NeurIPS Code of Ethics..

\item {\bf Broader impacts}
    \item[] Question: Does the paper discuss both potential positive societal impacts and negative societal impacts of the work performed?
    \item[] Answer: \answerYes{} 
    \item[] Justification: Please see Appendix.~\ref{impact}.
    
\item {\bf Safeguards}
    \item[] Question: Does the paper describe safeguards that have been put in place for responsible release of data or models that have a high risk for misuse (e.g., pre-trained language models, image generators, or scraped datasets)?
    \item[] Answer: \answerYes{} 
    \item[] Justification: We will properly give access to users with license only.

\item {\bf Licenses for existing assets}
    \item[] Question: Are the creators or original owners of assets (e.g., code, data, models), used in the paper, properly credited and are the license and terms of use explicitly mentioned and properly respected?
    \item[] Answer: \answerYes{} 
    \item[] Justification: All of the used datasets are properly credited

\item {\bf New assets}
    \item[] Question: Are new assets introduced in the paper well documented and is the documentation provided alongside the assets?
    \item[] Answer: \answerNA{} 
    \item[] Justification:

\item {\bf Crowdsourcing and research with human subjects}
    \item[] Question: For crowdsourcing experiments and research with human subjects, does the paper include the full text of instructions given to participants and screenshots, if applicable, as well as details about compensation (if any)? 
    \item[] Answer: \answerNA{}. 
    \item[] Justification:

\item {\bf Institutional review board (IRB) approvals or equivalent for research with human subjects}
    \item[] Question: Does the paper describe potential risks incurred by study participants, whether such risks were disclosed to the subjects, and whether Institutional Review Board (IRB) approvals (or an equivalent approval/review based on the requirements of your country or institution) were obtained?
    \item[] Answer: \answerYes{} 
    \item[] Justification: This paper only includes public datasets with proper approvals.

\item {\bf Declaration of LLM usage}
    \item[] Question: Does the paper describe the usage of LLMs if it is an important, original, or non-standard component of the core methods in this research? Note that if the LLM is used only for writing, editing, or formatting purposes and does \emph{not} impact the core methodology, scientific rigor, or originality of the research, declaration is not required.
    \item[] Answer: \answerNA{} 
    \item[] Justification: LLM is only used for polish writing in this research.

\end{enumerate}

\end{document}